%% file: main.tex
\pdfoutput=1

\documentclass[11pt]{article}

\usepackage[final]{acl}

\usepackage{times}
\usepackage{latexsym}

\usepackage[T1]{fontenc}

\usepackage[utf8]{inputenc}

\usepackage{microtype}

\usepackage{inconsolata}

\usepackage{graphicx}

\usepackage{amsmath}
\usepackage{amsfonts}

\usepackage{makecell}

\usepackage{float}
\usepackage{stfloats}

\usepackage{algorithm}
\usepackage{algpseudocode}

\algrenewcommand\Require{\item[\textbf{Input:}]}
\algrenewcommand\Ensure{\item[\textbf{Output:}]}

\usepackage{svg}

\usepackage{booktabs}
\usepackage{multirow}
\usepackage{xcolor}
\usepackage{colortbl}

\usepackage{gensymb}

\usepackage{hyperref}

\usepackage{fdsymbol}

\usepackage{soul}

\newcommand{\ourModel}{VEHME}

\title{\ourModel: A Vision-Language Model For Evaluating Handwritten Mathematics Expressions}

\author{
 \textbf{Thu Phuong Nguyen\textsuperscript{$\spadesuit$}\thanks{Both authors contributed equally to this work.}}\qquad
 \textbf{Duc M. Nguyen\textsuperscript{$\diamondsuit$}\footnotemark[1]} \\
 \textbf{Hyotaek Jeon\textsuperscript{$\diamondsuit$}}\qquad 
 \textbf{Hyunwook Lee\textsuperscript{$\diamondsuit$}}\qquad
 \textbf{Hyunmin Song\textsuperscript{$\spadesuit$}}\\
 \textbf{Sungahn Ko\textsuperscript{$\diamondsuit$}\thanks{Co-corresponding authors.}}\qquad
 \textbf{Taehwan Kim\textsuperscript{$\spadesuit$}\footnotemark[2]}
 \\
 \textsuperscript{$\spadesuit$}Ulsan National Institute of Science and Technology\\
 \textsuperscript{$\diamondsuit$}Pohang University of Science and Technology\\
 \texttt{\{phuongnt, hyunminsong, taehwankim\}@unist.ac.kr}\\
 \texttt{\{ducnm, taek98, hwlee0916, sungahn\}@postech.ac.kr}
}

\begin{document}
\maketitle
\input{section/0_Abstract}   
\input{section/1_Intro}

\input{section/2_Related}

\input{section/3_Method}

\input{section/4_Experiments}

\input{section/5_Conclusion}
\input{section/6_Limitations}
\input{section/7_Acknowledgments}

\bibliography{custom}
\input{section/X_Appendix}

\end{document}

%% file: section/0_Abstract.tex
\begin{abstract}
Automatically assessing handwritten mathematical solutions is an important problem in educational technology with practical applications, but it remains a significant challenge due to the diverse formats, unstructured layouts, and symbolic complexity of student work. To address this challenge, we introduce VEHME-a \textbf{V}ision-Language Model for \textbf{E}valuating \textbf{H}andwritten \textbf{M}athematics \textbf{E}xpressions—designed to assess open-form handwritten math responses with high accuracy and interpretable reasoning traces. VEHME integrates a two-phase training pipeline: (i) supervised fine-tuning using structured reasoning data, and (ii) reinforcement learning that aligns model outputs with multi-dimensional grading objectives, including correctness, reasoning depth, and error localization. To enhance spatial understanding, we propose an Expression-Aware Visual Prompting Module, trained on our synthesized multi-line math expressions dataset to robustly guide attention in visually heterogeneous inputs. Evaluated on AIHub and FERMAT datasets, VEHME achieves state-of-the-art performance among open-source models and approaches the accuracy of proprietary systems, demonstrating its potential as a scalable and accessible tool for automated math assessment. Our training and experiment code is publicly available at our \href{https://github.com/phuongnt197/VEHME}{GitHub repository}.
\end{abstract}

%% file: section/1_Intro.tex
\section{Introduction}

The assessment of handwritten mathematical solutions is a fundamental yet labor-intensive component of mathematics education, serving as both a diagnostic tool for educators and a critical feedback mechanism for learners \cite{Tutor, lin2024ai, zhang2021math, nakamoto2025development}. 
In traditional classroom settings, teachers manually evaluate student work to identify conceptual misunderstandings, procedural errors, and gaps in logical reasoning. 
But, the scalability of this process is severely limited by time constraints, class sizes, and the cognitive load of interpreting diverse solution strategies \cite{gowda2017comparative, laws2003scalability, callahan2021leveraging}. 
With the rise of digital learning platforms, there is an urgent need for automated systems that can replicate the nuanced judgment of human graders while accommodating the unstructured, creative, and often messy nature of student-written content \cite{DrawEduMath, Tutor}.

\begin{figure}[t]
    \centering
    \includegraphics[width=\linewidth]{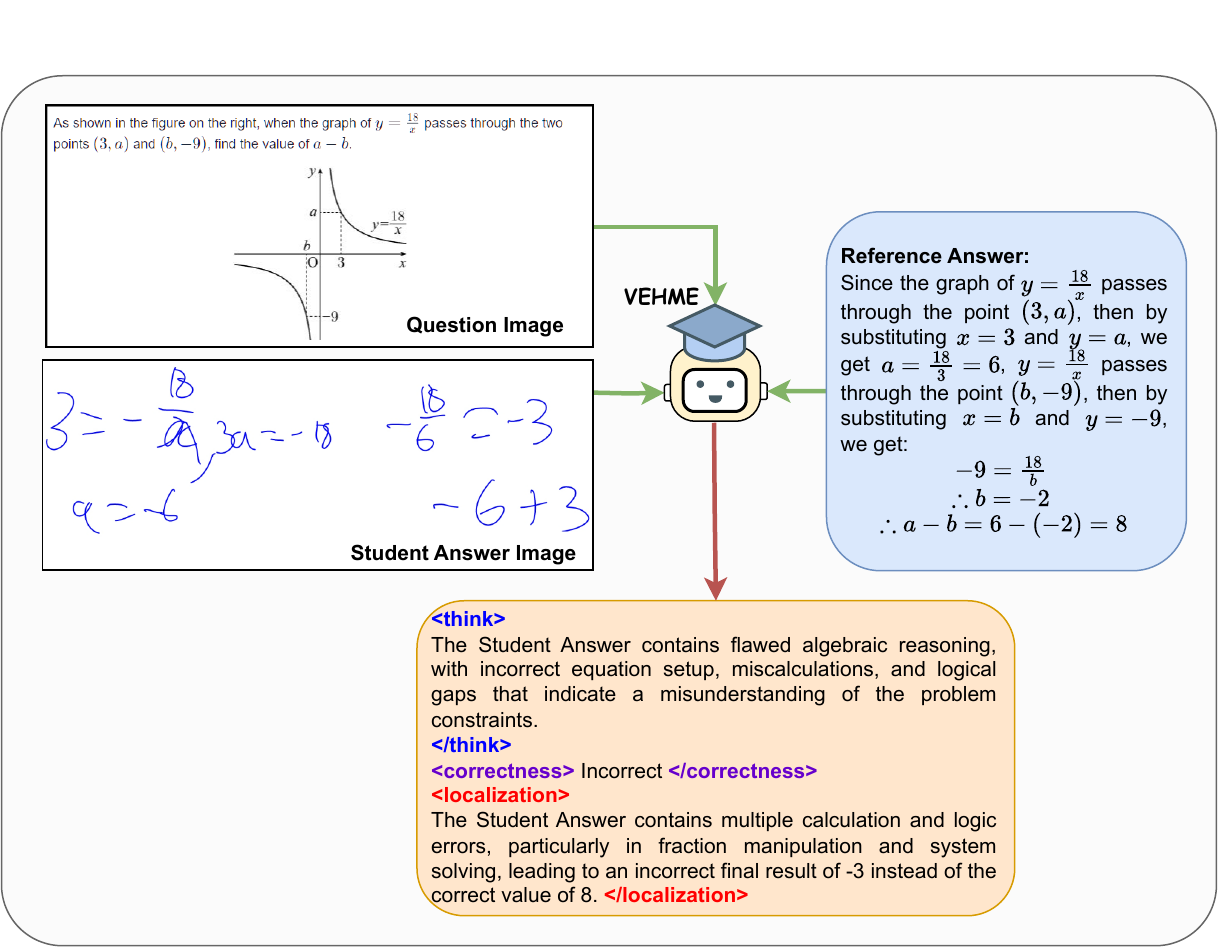}
    \caption{Our model takes a question, reference answer, and student answer image as input to predict the correctness of the student's solution and identify any error locations, if the solution is incorrect.}
    \label{fig:1}
\end{figure}

To address scalability challenges in grading, prior work has explored automated systems built around structured templates and Optical Character Recognition (OCR) pipelines \cite{Bags, HmwkCheck, MathMistakeChecker, wagstaff2019automatic}, as well as typed-text inputs processed by language models \cite{Tutor}. 
However, these approaches rely on rigid assumptions: that student responses are syntactically clean, spatially structured, or written within constrained templates. 
Furthermore, there is a notable lack of high-quality datasets that include complex, multi-expression handwritten math expressions \cite{yang2023read}, limiting the capacity to model open-form student solutions. 
Consequently, these systems often break down when applied to real-world data that is handwritten, unstructured, and heterogeneous, featuring symbolic math, diagrams, and natural language, all rendered in varied handwriting styles (Figure~\ref{fig:1}). 

Previous approaches that leverage Large Language Models (LLMs) for grading often depend solely on the model's end-to-end reasoning capabilities without explicitly supervising or constraining the underlying reasoning process \cite{MathMistakeChecker, phys, thermo}. 
In the text-only domain, Process Reward Models (PRMs) have been introduced to address this issue by enabling step-by-step evaluation of mathematical sequential reasoning \cite{skyworkopeno12024, math-shephred, lambert2024rewardbench}. 
More recently, PRMs have been extended into the multimodal domain \cite{khalifa2025processthink, wang2025visualprm}, allowing reasoning-aware supervision in visual tasks. 
However, to the best of our knowledge, there is still no effective approach tailored to handwritten mathematical expressions, a domain that combines the complexities of symbolic reasoning with the challenges of visual noise and layout diversity.

While recent advances in Vision-Language Models (VLMs) have shown promise in multimodal understanding, their application to handwritten math assessment has been limited by several key challenges: (C1) the scarcity of high-quality training data that pairs handwritten expressions with detailed error annotations \cite{MathCog}, (C2) the difficulty in robustly processing the spatial layout and visual noise characteristic of student work \cite{vlpg}.


In this work, we take a step toward overcoming these challenges by tackling the underexplored and difficult task of grading template-free, handwritten open-response solutions in the broader domain of K–12 mathematics. 
This setting requires models capable of handling significant visual heterogeneity, unstructured layouts, and sequential reasoning, where early mistakes may propagate across later steps. 
To this end, we present \textbf{\ourModel} (a \textbf{V}ision-Language Model for \textbf{E}valuating \textbf{H}andwritten \textbf{M}athematics \textbf{E}xpressions), a novel end-to-end framework for automatically grading handwritten math solutions. 
Our key contributions are summarized as follows: 

\begin{itemize}

    \item To address the challenge of assessing complex open-form mathematical expressions, we propose a novel framework named \textbf{\ourModel}. To the best of our knowledge, we are the first to propose an end-to-end training pipeline for grading handwritten mathematical expressions.

    \item We adopt a dual-phase training regime: supervised fine-tuning on our synthesized datasets, followed by reinforcement learning with a composite reward to optimize for correctness, reasoning quality, and error localization. 
    
    \item To address C1, we also propose a data synthesis pipeline to construct a high-quality reasoning-augmented dataset that pairs handwritten expressions with error detection and localization reasoning traces. 
 
    
    \item To address C2, we propose the Expression-aware Visual Prompting Module (EVPM), equipping the VLM with precise spatial localization of handwritten mathematical expressions. To train this module, we construct a complex, heterogeneous, multi-expression handwritten math expressions dataset with its corresponding spatial information.
 
    \item Our training pipeline enables small, open-source Vision-Language Models (VLMs) to surpass the performance of larger open-source counterparts and achieve results on par with state-of-the-art proprietary models.
 
   \end{itemize}

%% file: section/2_Related.tex
\section{Related Work}
\subsection{Handwritten Mathematical Expression Recognition}
Handwritten Mathematical Expression Recognition (HMER) has been a long-standing and evolving research area. The inherent ambiguity and complex structure in handwritten symbols present significant challenges. Traditional methods typically employed a two-stage process: mathematical symbol recognition followed by structure analysis~\cite{chan1998elastic, chan2000mathematical}.

The advent of deep learning spurred a shift towards end-to-end OCR-based models. These approaches aimed to directly predict the symbolic representation (e.g., LaTeX) from an input handwritten mathematical expression image. Variations included models that directly output the sequence~\cite{deng2017image, li2022counting, wu2020handwritten, zhao2022comer} and others that incorporated grammatical rules of mathematical expressions to guide and constrain the prediction process~\cite{li2024tree, wu2022tdv2, wu2021graph}.

More recently, Vision Language Models (VLMs), which combine vision encoders with Large Language Models (LLMs), have shown strong multimodal capabilities~\cite{liu2023visual, bai2023qwenvlversatilevisionlanguagemodel, chen2024internvl}. There is emerging interest in applying VLMs to HMER, leveraging their potential for understanding complex visual and textual relationships~\cite{hievl}.

\subsection{Automatic Grading of Handwritten Math}
The labor-intensive nature of manually grading handwritten mathematical solutions has driven the demand for automated systems. Early efforts in automated grading were often limited to structured or short-answer questions due to technological constraints. A prevalent traditional method involved converting handwritten expressions to a format like LaTeX using the HMER model, followed by evaluation with rule-based system or heuristics~\cite{Bags, HmwkCheck, chaowicharat2023step}.

With the development of LLMs, new approaches have emerged where LLMs compare a student's formatted submission (derived via HMER) against a correct solution~\cite{MathMistakeChecker}. 
For finer evaluation (e.g., assessing complex solutions or providing step-level feedback), reinforcement learning (RL) has been explored to facilitate error localization and step-by-step correction~\cite{StepAMC, processbench}.
VLMs have also been investigated for direct automatic grading from handwritten input, aiming to bypass the explicit HMER formatting stage~\cite{fermath, DrawEduMath, MathCog, phys}.
However, the VLM-based systems show considerable sensitivity to handwriting quality, which currently restricts their robustness in practical, real-world scenarios.

\subsection{Mathematical Reasoning and LLM as a Judge}
The recent success of reasoning-focused models across domains such as mathematics and programming \cite{openai2024reasoning} draws significant attention to enhancing the reasoning capabilities of LLMs through reinforcement learning (RL) \cite{zhang2024o1, zhang2024codedpo, trung2024reft, chen2024alphamath}. 
In this context, \citet{deepseek-r1} introduced a novel RL-based approach that removes the need for supervised fine-tuning while achieving strong and reliable reasoning performance. Building on this progress, recent efforts have begun exploring similar techniques in multi-modal settings, attracting increasing research interest \cite{Liu2025VisualRFTVR, Deng2025OpenVLThinkerAE, Zhou2025R1ZerosM}.

In tasks requiring open-ended responses, such as error localization in educational contexts or code reviews, human evaluation is often necessary to validate correctness. However, this dependence on manual judgment limits scalability. Consequently, there is growing interest in using LLMs themselves as automated judges to evaluate the quality and accuracy of responses \cite{zheng2023judging, chen2024mllm, gu2024survey}. Recent work has started to integrate LLM-based evaluators directly into RL training pipelines for such open-ended tasks.

%% file: section/3_Method.tex
\section{Methodology}

\begin{figure*}[t]
  \includegraphics[width=\linewidth]{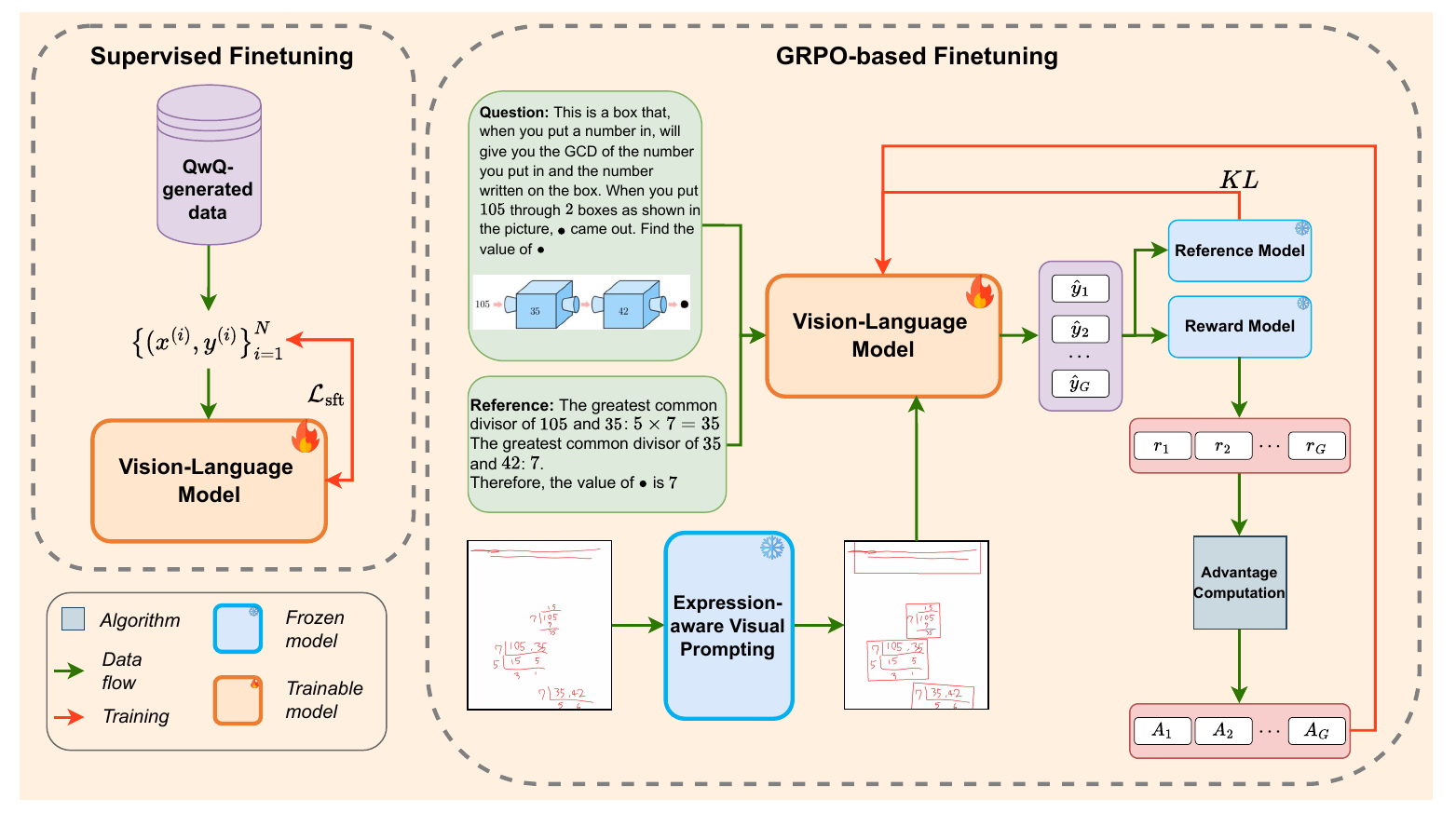}
  \caption {\ourModel\ overview. Initially, a VLM is fine-tuned to generate outputs in the desired format using synthesized data from Sec~\ref{sec:data-synthesis}. A subsequent Preference Optimization step trains the model using GRPO~\cite{deepseekmath} method. This optimization is guided by rewards described in Sec~\ref{sec:reward-modeling} and takes the given problem, reference answer, and an expression-aware visual-prompted image from the student's answer as input. 
  }
  \label{fig:pipeline}
\end{figure*}

Our approach to Handwritten Mathematics Grading integrates two key components: (1) supervised fine-tuning using distilled data from the \verb|QwQ-32B| model to instill foundational reasoning capabilities, and (2) reinforcement learning via Group Relative Policy Optimization \cite{deepseekmath} to refine the model's ability to generate accurate and explainable grading outputs. As shown in the Figure~\ref{fig:pipeline}, this dual-phase training strategy aims to enhance both the correctness assessment and the interpretability of the model's evaluations.

\subsection{Problem Formulation}
We formulate the \emph{Handwritten Mathematics Grading} problem as a vision-language modeling task, where the goal is to assess the correctness of a student's handwritten mathematical answer based on a given question and a reference solution.

Let each instance be represented by the tuple \( x = (q, r, s) \), where:
(i) \( q \in \mathbb{P}^* \) is the \textbf{question}, represented as a sequence of pixels (e.g., an image of a printed or handwritten mathematical question),
(ii) \( r \in \mathbb{V}^* \) is the \textbf{reference answer}, expressed as a sequence of structured mathematical tokens from a vocabulary \( \mathbb{V} \), and
(iii) \( s \in \mathbb{P}^* \) is the \textbf{student answer}, also represented as a sequence of pixels (i.e., a handwritten response).
Here, \( \mathbb{P} \) denotes the space of pixel values (e.g., grayscale or RGB), and \( \mathbb{V} \) denotes the vocabulary of natural language and symbolic mathematical expressions (e.g., LaTeX or semantic tokens).
Each instance is associated with a grading label \( y = (c, l) \), where:
(i) \( c \in \{0, 1\} \) indicates the \textbf{correctness} of the student answer, and
(ii) \( l \in \mathbb{V}^* \) denotes the \textbf{location or description of the error}, provided as a (possibly empty) subsequence of tokens aligned to the student answer.

Let \(\mathbb{X} = \mathbb{P}^* \times \mathbb{V}^* \times \mathbb{P}^*\) and \(\mathbb{Y} = \{0, 1\} \times \mathbb{V}^*\) denote the input and output spaces, respectively. The goal is to learn a function
\(
f: \mathbb{X} \rightarrow \mathbb{Y},
\)
that maps each input tuple \( x = (q, r, s) \) to a grading label \( y = (c, l) \), capturing both correctness and, if applicable, the symbolic location of the error in the student's response.
To approximate \( f \), we train a vision-language model $\pi_{\theta}$
parameterized by \( \theta \), which defines an autoregressive conditional distribution \( \pi_{\theta}(y_t \mid x, y_{<t}) \) over tokens in the output space. The function \( f \) is then approximated by
\(
f(x) \approx \arg\max_{y \in \mathbb{Y}^{*}} \prod_{t=1}^{T} \pi_{\theta}(y_t \mid x, y_{<t}),
\)
where \( T \) is the length of the output sequence \( y \), and \( y_{<t} \) denotes the previously generated tokens.

\subsection{Main Techniques}

\subsubsection{Data Synthesis and Supervised Fine-tuning}
\paragraph{Data synthesis from QwQ-32B}
\label{sec:data-synthesis}

\begin{figure*}[t]
  \includegraphics[width=\linewidth]{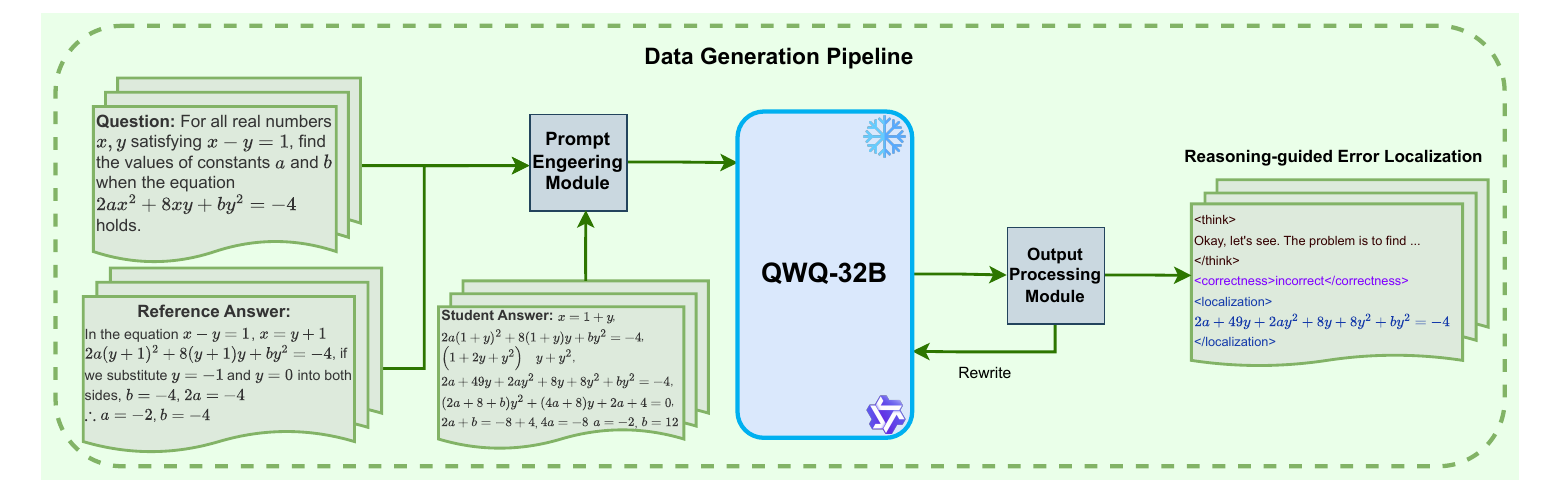}
  \caption {Process for synthesizing SFT data. \texttt{QwQ-32B} creates structured feedback from inputs \textit{(q, r, s)} under a token budget \textit{M}. Truncated outputs are repaired using a output processing module (Appendix~\ref{sec:data-syn}) with grammar-constrained decoding (Appendix~\ref{sec:data-syn}) to ensure valid SFT data.}
  \label{fig:sft_data}
  \vspace{-0.3cm}
\end{figure*}


\citet{processbench} demonstrates that \verb|QwQ-32B| \cite{qwq32b}, a large language model built upon Qwen2.5-32B \cite{qwen2.5} and optimized specifically for mathematical reasoning, achieves state-of-the-art performance among open-source LLMs on step-level mathematical evaluation tasks, even surpassing fine-tuned process reward models (PRMs). 
This strong reasoning capability, despite \verb|QwQ-32B|'s relatively modest size compared to models like GPT-4 \cite{gpt4} or DeepSeek-R1 \cite{deepseek-r1}, makes it a compelling teacher model for mathematical evaluation.

To capitalize on this, we perform supervised fine-tuning using data distilled from \verb|QwQ-32B| as shown in the Figure~\ref{fig:sft_data}. 
Given a tuple \( x = (q, r, s) \) consisting of a question, a reference answer, and a student's handwritten response, we prompt \verb|QwQ-32B| to assess the correctness of the answer and, if incorrect, to identify or describe the error. The model is instructed to generate responses in a structured format composed of three fields: a detailed thinking process, a correctness label, and an error localization. We set a maximum token budget of \(M\) for efficient training, as longer models' responses do not necessarily improve the model's performance \cite{chen2025judgelrm, yeo2025demystifying, Deng2025OpenVLThinkerAE, Chen2025SFTOR}.
In cases where the model is truncated due to exceeding the token budget, we apply a structured post-processing routine that repairs truncated outputs and ensures conformance to the pre-defined format. Specifically, we truncate the response after the last complete thought from the model, append a closing \texttt{"</think>"} tag, and resume generation starting from \texttt{"<correctness>"} using grammar-constrained decoding \cite{geng2023grammarconstrained}. This allows the remainder of the response to be completed in a syntactically valid manner. The full post-processing procedure, the grammar, and more samples of the synthesized dataset can be found in Appendix~\ref{sec:data-syn}.

\paragraph{Supervised Fine-Tuning} 
The distilled dataset from \verb|QwQ-32B| provides high-quality demonstrations of expert-level reasoning and error localization. We use this data to initialize our model via supervised fine-tuning (SFT) (Figure~\ref{fig:pipeline}), training it to mimic the teacher model’s structured responses through next-token prediction.

\subsubsection{Reinforcement Learning with Group Relative Preference Optimization}

To improve the model's ability to produce accurate and explainable grading outputs, we adopt Group Relative Policy Optimization (GRPO) \cite{deepseekmath}, a reinforcement learning technique that leverages intra-group comparisons among multiple completions generated per prompt, effectively optimizing the relative ranking of responses without relying on value function estimation. This approach has demonstrated significant improvements in mathematical reasoning tasks.

Recent studies have shown that GRPO facilitates stable training and emergent reasoning behavior, often surpassing supervised fine-tuning \cite{Chen2025SFTOR, Deng2025OpenVLThinkerAE, Zhou2025R1ZerosM, Shen2025VLMR1AS}. \citet{Liu2025VisualRFTVR} further emphasizes GRPO’s robustness, while \citet{chen2025judgelrm} shows that judgment tasks, closely related to our grading tasks, are reasoning-intensive and benefit significantly from GRPO-based optimization.

\paragraph{GRPO formulation} Formally, for each input \( x \), we sample a group \( G = \{\hat{y}_1, \hat{y}_2, \ldots, \hat{y}_{|G|}\} \) of completions from the current policy \( \pi_\theta \). Each response \( \hat{y}_i = (\hat{c}_i, \hat{l}_i) \) is evaluated using a composite function \( r(\hat{y}_i) \). The advantage is efficiently estimated using the standardization of the rewards within the group:
\(
\hat{A}_i = \frac{r(\hat{y}_i) - \mu_G}{\sigma_G},
\) 
where \( \mu_G \) and \( \sigma_G \) are the mean and standard deviation of the rewards in group \( G \), respectively.
The GRPO objective is then formulated similarly to \citet{deepseekmath} as:
\begin{align*}
    &\mathcal{J}_\text{GRPO} = \mathbb{E}[x \sim P(X), \{y_i\}_{i=1}^{|G|} \sim \pi_{\theta_\text{old}}\left(y_i|x\right)] \\
    &\dfrac{1}{|G|}\sum_{i=1}^{|G|}\dfrac{1}{|\hat{y}_i|}\sum_{t=1}^{|\hat{y}_i|}{\left[\min{\left(\hat{A}_i\rho_{i, t}, \hat{A}_i\varsigma_{i, t}\right)} - \beta \mathcal{D}_\theta\right]},\\
    &\varsigma_{i, t} = \text{clip}{\left(\rho_{i, t}, 1 - \epsilon, 1 + \epsilon\right)},\\
    &\mathcal{D}_\theta = \mathbb{D}_{KL}{\left[\pi_\theta\mid\mid\pi_\text{ref}\right]},
\end{align*}
where \( \rho_{i, t} = \frac{\pi_\theta(\hat{y}_{i, t} \mid x, \hat{y}_{i, <t})}{\pi_{\theta_{\text{old}}}(\hat{y}_{i, t} \mid x, \hat{y}_{i, <t})} \) is the probability ratio between the current and previous policies, \( \mathbb{D}_{KL}{\left[\pi_\theta||\pi_\text{ref}\right]} \) is the unbiased estimator of the KL divergence between the current and the reference policies \cite{schulman2017proximal}, and \( \epsilon, \beta \) are hyperparameters controlling the clipping range and the KL penalty \cite{deepseekmath}. \

\paragraph{Reward Modeling}
\label{sec:reward-modeling}

Each response is scored using a composite reward function:
\begin{align*}
    r(\hat{y}_i) &= r_{\text{match}}(\hat{y}_i) + r_{\text{loc}}(\hat{y}_i) + r_{\text{len}}(\hat{y}_i)\\
                 &+ r_{\text{cos}}(\hat{y}_i) + r_{\text{rep}}(\hat{y}_i), 
\end{align*}
where
\begin{align*}
r_{\text{match}}(\hat{y}_i) &= \begin{cases}
    1 & \text{if}\ \hat{c}_i = c_i\\
    0 & \text{otherwise}
\end{cases},\\
r_{\text{loc}}(\hat{y}_i) &= \begin{cases}
    1 & \text{if}\ \hat{l}_i\ \text{correctly describe the error}\\
    0 & \text{otherwise}
\end{cases},\\
r_{\text{len}}(\hat{y}_i) &= \begin{cases}
    0.25 & \text{if}\ |\hat{y}_i| \geq 150\\
    0 & \text{otherwise}
\end{cases},
\end{align*}
and \( r_{\text{cos}}, r_{\text{rep}} \) are the Cosine Reward and the repetition penalty \cite{yeo2025demystifying}. For \(r_\text{loc}\), traditional rule-based evaluation is inadequate due to the open-ended nature of the task and the variability in valid explanations. Prior work has demonstrated that QWQ-32B achieves great results in text-based step-level error localization \cite{processbench}. Thus, we opt to use QWQ-32B as an automated judge to evaluate the correctness and quality of the model's error localization outputs for this component of the reward function.

\subsubsection{Expression-aware Visual Prompting}

\begin{figure}[t]
  \includegraphics[width=1\linewidth]{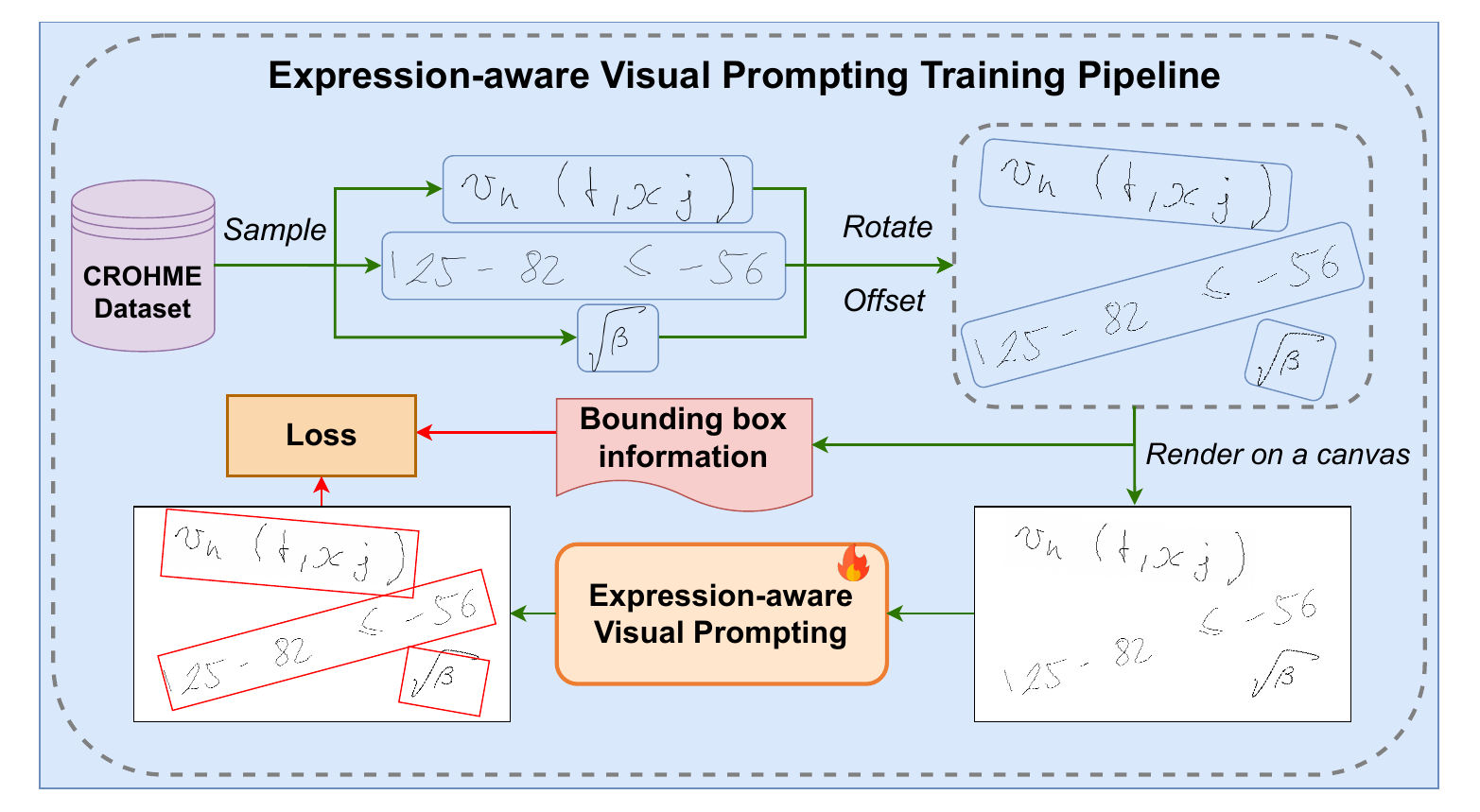}
  \caption{EVPM training pipeline. The model learns to predict ground truth bounding boxes for augmented mathematical expressions which are drawn onto a white backgrounds after augmentation.} 
  \label{fig:GenBBox}
  \vspace{-0.5cm}
   
\end{figure}

Previous studies have shown that Vision-Language Models (VLMs) struggle to interpret multi-line handwritten mathematical expressions, which are common in students' solutions \cite{hievl}. Recognizing these complex layouts is essential for downstream tasks such as automated grading. To address this, we propose an \textit{Expression-aware Visual Prompting Module} (EVPM), which provides the model with spatial cues by generating oriented bounding boxes around individual mathematical expressions \cite{Li2024VisualPrompt}. 

The EVPM comprises an oriented bounding box predictor trained on a synthetic dataset of multi-line mathematical expressions (Figure~\ref{fig:GenBBox}). To train the EVPM component, we utilize Yolov11 \cite{Jocher_Ultralytics_YOLO_2023} as the backbone of the bounding box detector. We construct the synthetic dataset by first sampling expressions from the grammar-based generator proposed by \citet{truong2022syntactic}, then rendering each expression as a handwritten image using symbol-level traces. Recognizing that students often write in non-linear or skewed patterns, particularly when unbounded by ruled lines, we simulate these realistic distortions through a stochastic layout pipeline. Each expression is randomly rotated and placed on a blank canvas with vertical spacing and padding variation, forming a more naturalistic setting for bounding box supervision. To create such expression-rich canvases, we formalize the image construction process in Appendix~\ref{sec:evp}. Each mathematical expression is first rendered into an image, then randomly rotated and positioned vertically on the canvas. The corresponding oriented bounding boxes are computed by rotating each image's axis-aligned box by the same angle. This process supports diverse and realistic handwriting layouts during training.

By training the EVPM on these synthetic examples, the model learns to identify individual expressions even under irregular layouts. This bounding box information is provided to the downstream VLM as visual prompts, enabling more accurate understanding of complex student solutions.

%% file: section/4_Experiments.tex
\section{Experiments}

\subsection{Evaluation Datasets and Metrics}
\label{sec:dataset}
We use two datasets for handwritten mathematical assessment: the AIHub dataset\footnote{\href{https://aihub.or.kr/aihubdata/data/view.do?currMenu=&topMenu=&aihubDataSe=data&dataSetSn=71716}{https://aihub.or.kr}} and the FERMAT dataset \cite{fermath}. More information about datasets' statistics and samples can be found in Appendix~\ref{sec:datasets}.

\paragraph{AIHub dataset} is derived from a large-scale educational repository encompassing K–12 mathematics problems. 
It comprises 183,085 handwritten student responses corresponding to 30,050 unique problem questions, originally authored in Korean. 
To mitigate potential performance degradation due to cross-lingual inputs, we filtered out student answers containing non-English characters. 
The remaining questions and reference answers were translated into English using the Google Translate API\footnote{\href{https://cloud.google.com/translate/docs/reference/rest}{https://cloud.google.com/translate/docs/reference/rest}}. 
In the end, we obtained 81,394 training samples and 9,062 test samples for the AIHub dataset.

\paragraph{FERMAT dataset} provides a complementary evaluation focused on error analysis, so incorrect solutions significantly outnumber correct ones at an approximate ratio of 85:15, which is different from the balanced distribution in the filtered AIHub dataset. It contains 2,244 manually curated solutions across eight mathematical domains from arithmetic to calculus. Each entry contains: the original question, a handwritten solution image with intentional errors, the gold-standard correct answer, and error detection labels. Due to its small size, we opt not to perform any additional filtering, as doing so could lead to an insufficient amount of data for fine-tuning. We split the dataset into 70\% training and 30\% test sets.

\paragraph{Metrics}
For error detection performance evaluation, we employ two key metrics-- 
\textbf{Accuracy}: measures overall prediction correctness across both error-present and error-free samples, and
\textbf{F1-Score}: balances precision and recall to handle class imbalance, particularly crucial for FERMAT's skewed error distribution.

\subsection{Implementation Details}
We conduct our experiments using Qwen2.5-VL-7B-Instruct \cite{qwen-vl}, a pre-trained large multimodal model. 
As an initial step, we perform supervised fine-tuning (SFT) using text-only data distilled from \verb|QwQ-32B| \cite{qwq32b} to align the model with the desired output format. During reinforcement learning with GRPO, input images are resized to limit visual token overhead, capped at 50,176 pixels (224×224 image resolution) for the AIHub dataset and 501,760 pixels for FERMAT, preserving essential content while improving training efficiency. For more details, please refer to Appendix~\ref{sec:appendix-implementation-details}.

\subsection{Comparison with the State-of-the-Art}

\paragraph{Experimental setup} 

\begin{figure}[t]
    \centering
  \includegraphics[width=0.85\linewidth]{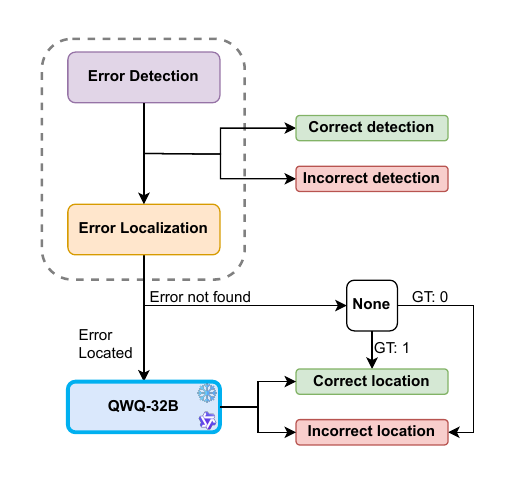}
  \vspace{-0.3cm}
  \caption {VLM evaluation procedure for student solutions, encompassing Error Detection (ED) and Error Localization (EL). ED serves to identify the presence of errors, while EL is responsible for determining the specific location of any detected error.} 
  \label{fig:inference}
  \vspace{-0.3cm}
\end{figure}

We conduct experiments on two subtasks of the mathematical grading task: Error Detection (ED) and Error Localization (EL), evaluating a cascaded setup where ED and EL are executed sequentially (Figure~\ref{fig:inference}). 
The input for VLM processing consists of a problem question, a reference answer, and a student's solution.
For the ED task, the model outputs a binary classification indicating whether the student's solution contains an error (Incorrect solution) or is error-free. 
Performance is evaluated based on the model's classification accuracy. 
If the student's answer contains an error, the model proceeds to identify the specific location of the error for the EL task. 
To assess localization precision, we use the critic model QwQ-32B as a Judge due to its superior performance in error localization \cite{processbench}. 
When no error location is identified (None), we cross-check with the ground truth of error detection—if the student's answer is correct, the localization is deemed correct; otherwise, it is marked as wrong localization.
The results are reported based on a single run for consistency. 
Further details on datasets (e.g., AIHub, Fermat) and evaluation protocols are provided in the Appendix.

\paragraph{Experimental results } 

\newcommand{\vwidth}{0.75pt}
\begin{table*}[t]
\centering
\renewcommand{\arraystretch}{0.8}
\begin{tabular}{
    l
    !{\vrule width \vwidth}
    cc|cc
    !{\vrule width \vwidth}
    cc|cc
}
\toprule
\multirow{2}{*}{\textbf{Models}} & \multicolumn{4}{c!{\vrule width \vwidth}}{\textbf{AIHub}} & \multicolumn{4}{c}{\textbf{FERMAT}} \\
\cmidrule(lr){2-9}
& \multicolumn{2}{c|}{\textbf{ED}} & \multicolumn{2}{c!{\vrule width \vwidth}}{\textbf{EL}} & \multicolumn{2}{c|}{\textbf{ED}} & \multicolumn{2}{c}{\textbf{EL}} \\
& Acc & F1 & Acc & F1 & Acc & F1 & Acc & F1 \\
\midrule
\rowcolor{gray!20} \multicolumn{9}{c}{Closed-source Models} \\
 \midrule

\rowcolor{gray!20} GPT-4o & 73.00 & 48.67 & 62.88 & 61.01 & 66.62 & \textbf{58.57} & 55.04 & 66.07  \\
\rowcolor{gray!20} GPT-4o-mini & 57.15 & 32.97 & 38.98 & 49.18 & \textbf{81.75} & 56.62 & 47.18 & 62.61  \\
\rowcolor{gray!20} Gemini-2.0-Flash & \textbf{75.40} & \textbf{50.30} & \textbf{67.40} & \textbf{65.73} & 75.22 & 43.13 & \textbf{64.24} & \textbf{75.13}  \\
\rowcolor{gray!20} Gemini-2.5-Flash-Preview & 71.01 & 46.91 & 63.00 & 65.68 & 80.12 & 45.68 & 63.35 & 74.72  \\
\midrule
\multicolumn{9}{c}{Open-source Models} \\
\midrule
Phi-4-multimodal-instruct & 38.57 & 29.39 & 35.93 & 21.33 & 36.20 & 22.33 & 16.91 & 15.15  \\
Qwen2.5-VL-7B-Instruct & 46.68 & 31.48 & 38.00 & 33.75 & 44.81 & 26.04 & 24.33 & 32.18  \\
Pixtral-12B & 52.67 & 31.69 & 32.20 & 38.35 & 41.39 & 25.56 & 10.53 & 3.52  \\
Llama-3.2-11B-Vision-Instruct & 19.49 & 18.58 & 44.78 & 5.62 & 32.49 & 22.15 & 21.07 & 22.67  \\

\midrule

\ourModel-Qwen2.5-VL-7B & \textbf{73.01} &\textbf{ 49.22} &\textbf{ 61.13} & \textbf{ 58.18} & \textbf{62.61} & \textbf{29.81} & \textbf{31.90} & \textbf{44.36}  \\

\bottomrule
\end{tabular}
\caption{Performance comparisons of state-of-the-art Vision-Language Models on two datasets: AIHub and FERMAT \cite{fermath}. The evaluation metrics include Accuracy (Acc) and F1 score (F1). \textbf{ED}: Error Detection, \textbf{EL}: Error Localization. All of the reported results are in percentages ($\%$).}
\label{tab:model_performance}
\end{table*}

The experiments are conducted on two tasks--Error Detection and Error Localization, across the AIHub and FERMAT datasets.
Table~\ref{tab:model_performance} compares our \ourModel‐Qwen2.5‐VL‐7B against four open‑source baselines and four closed‑source systems on the AIHub and FERMAT benchmarks. 
Among the open‑source group, our model achieves a clear lead in both Error Detection (ED) and Error Localization (EL). 
On AIHub, \ourModel\ records 73.01\% ED accuracy (49.22\% F1) and 61.13\% EL accuracy (58.18\% F1), substantially outperforming Qwen2.5‑VL‑7B‑Instruct (46.68\%Acc/ 31.48\% F1 for ED, 38.00\% Acc/ 33.75\% F1 for EL), Pixtral‑12B (52.67\% / 31.69\% ED, 32.20\% / 38.35\% EL), Phi‑4‑multimodal‑instruct, and Llama‑3.2‑11B. 
This margin highlights the effectiveness of our data synthesis pipeline and dual‐phase training in refining both expression recognition and precise localization.

On FERMAT, open‑source models generally struggle, but our approach again leads the pack with 62.61\% ED accuracy (29.81\% F1) and 31.90\% EL accuracy (44.36\% F1). 
Notably, Pixtral‑12B exhibits an unusual split—high ED accuracy (41.39\%) yet very low EL performance (10.53\% Acc/3.52\% F1)--which echoes the observations in the original FERMAT study: its format alignment and output consistency are suboptimal for localization tasks \citep{fermath}. 
In contrast, our model’s balanced gains across both metrics demonstrate robust adherence to the expected answer format and reliable spatial grounding.

When we turn to closed‑source systems, Gemini‑2.0‑Flash remains the top performer overall (75.40\%/50.30\% ED, 67.40\%/65.73\% EL on AIHub; 75.22\%/43.13\% ED, 64.24\%/75.13\% EL on FERMAT), followed closely by GPT‑4o and its mini variant. 
These proprietary models, however, exceed tens or even hundreds of billions of parameters, whereas our Qwen2.5‐VL‐7B backbone contains only 7 billion parameters. 
The result demonstrates that our lightweight model surpasses other open‐source approaches and nearly matches closed-source models, while operating at a fraction of the scale of closed systems, underscoring the efficiency gains enabled by our targeted data generation and training regimen. The qualitative result and analysis are presented in Appendix~\ref{sec:appendix-qualitative}. Addtionally, we provide analysis of experimental result across different educational levels in Appendix~\ref{sec:appendix-exp-education}. 

\subsection{Ablation Study}

\begin{table}
    \centering
    \renewcommand{\arraystretch}{0.8}
  \begin{tabular}{l!{\vrule width \vwidth}cc!{\vrule width \vwidth}cc}
    \toprule
    \multirow{2}{*}{\textbf{Models}} & \multicolumn{2}{c!{\vrule width \vwidth}}{\textbf{ED}} & \multicolumn{2}{c}{\textbf{EL}} \\
    \cmidrule(lr){2-5}
    & Acc & F1 & Acc & F1 \\
    \midrule
    Ours & \textbf{73.01} & \textbf{49.22} & \underline{61.13} & \textbf{58.18} \\
    \small -EVPM & \underline{71.98} & \underline{48.87} & \textbf{61.70} & \underline{56.59} \\
    \small -SFT & 63.24 & 41.47 & 52.67 & 33.41 \\
    \small -RL & 46.90 & 33.82 & 39.24 & 29.59 \\
    \midrule
    Baseline & 46.68 & 31.48 & 38.00 & 33.75 \\
    \bottomrule
  \end{tabular}
  \caption{\label{tab:ablation}
    Ablation study results validating our proposed components on the AIHub dataset.
  }
  \vspace{-0.4cm}
\end{table}

\begin{table}[tb]
    \centering
    \renewcommand{\arraystretch}{0.8}
  \begin{tabular}{l!{\vrule width \vwidth}cc!{\vrule width \vwidth}cc}
    \toprule
    \multirow{2}{*}{\textbf{Models}} & \multicolumn{2}{c!{\vrule width \vwidth}}{\textbf{ED}} & \multicolumn{2}{c}{\textbf{EL}} \\
    \cmidrule(lr){2-5}
    & Acc & F1 & Acc & F1 \\
    \midrule
    \small GPT-4o & \underline{73.00} & \underline{72.93} & \underline{63.00} & \underline{62.36} \\
    \small GPT-4o-mini & 48.00 & 39.22 & 33.00 & 29.26 \\
    \small Gemini-2.0-Flash & {71.00} & {71.00} & \textbf{65.00} & \textbf{64.83} \\
    \small Gemini-2.5-Flash* & \textbf{76.00} & \textbf{74.78} & {59.00} & {58.80} \\
    \midrule
    \small Phi-4* & 33.00 & 25.14 & 36.00 & 33.33 \\
    \small Qwen2.5-VL* & 50.00 & 35.87 & 46.00 & 45.45 \\
    \small Pixtral* & 41.00 & 25.38 & 29.00 & 28.82 \\
    \small Llama-3.2* & 21.00 & 18.93 & 48.00 & 34.04 \\
    \midrule
    Ours & \textbf{75.00} & \textbf{50.24} & \textbf{66.00} & \textbf{65.32} \\
    \small -EVPM & \underline{62.00} & \underline{42.26} & \underline{62.00} & \underline{61.00} \\
    \small -SFT & 61.00 & 39.02 & 58.00 & 53.21 \\
    \small -RL & 43.00 & 31.69 & 44.00 & 41.67 \\
    \bottomrule
  \end{tabular}
  \caption{\label{tab:ablation_challenging}
    Ablation study results validating our proposed components on the ``\textit{Challenging Subset of AIHub}'' dataset. *We use the same models as in the main experiment; we omit the suffix for brevity.}
  
  \vspace{-0.4cm}
\end{table}

\paragraph{Component Ablation} In this section, we perform a comprehensive ablation study on the balanced AIHub dataset to quantify the impact of each major component in our architecture. 
Table~\ref {tab:ablation} reports performance when ablating specific modules from our full model. 
Omitting the bounding‑box visual hints (–EVPM) yields a modest decrease in ED accuracy (73.01\%→71.98\%) and a larger decline in EL F1 (58.18\%→56.59\%), underscoring the necessity of spatial grounding. 
Skipping supervised fine‑tuning (–SFT) substantially degrades both ED and EL, indicating that task‑specific supervision is crucial for aligning the vision–language backbone with the reasoning steps and expected output format. 
The removal of reinforcement learning (–RL) results in a dramatic collapse, particularly in EL F1 (58.18\%→29.59\%), highlighting RL’s indispensable role in honing expression comprehension and precise localization. 
Our intact model consistently outperforms all ablated variants across every metric in both tasks.

\paragraph{Ablation Under Heavy Rotations} In our analysis, most of the standard dataset has mild rotation $(< 15\degree)$, which might explain the relatively small global improvement from EVPM. To better understand EVPM’s contribution, we constructed a ``\textit{Challenging Subset of AIHub}'' by identifying the 100 handwritten data points with the most heavily rotated mathematical expressions (top-100; Mean: $21.81\degree$). Examples of this subset can be found in Appendix~\ref{sec:challenging}. Table~\ref{tab:ablation_challenging} reports performance when the ablation study is conducted in the challenging subset. These results demonstrate that EVPM makes a significant contribution under heavy rotations. On this subset, removing EVPM causes a substantial drop in performance (ED: $75\% \to 62\%$, and EL: $66\% \to 62$), even with all other training and inference settings held constant. 
Compared with the closed-source models, our approach significantly narrows the gap in error detection and even outperforms them in error localization. Specifically, for ED, our method (75\%/50.24\%) comes close to the strongest closed-source baselines such as GPT-4o (73\%/72.93\%) and Gemini-2.5-Flash-Preview (76\%/74.78\%), showing that our system can achieve comparable detection accuracy under challenging rotations. More notably, in EL, our approach achieves 66\% Acc/65.32\% F1, surpassing both GPT-4o (63\%/62.36\%) and Gemini-2.5-Flash-Preview (59\%/58.80\%). This indicates that while the closed-source models remain competitive in ED, our method demonstrates a clear advantage in localizing errors under heavy rotational noise.

\paragraph{Image Resolution Ablation} In our main experiment, we resize the input images down to 224x224 for computational efficiency. To assess the impact of image resolution on performance, we conduct an ablation study using images with varying resolutions. Table~\ref{tab:ablation_image_size} shows \ourModel's performance across resolutions. Increasing the resolution generally improves both error detection and localization accuracies, with the largest gains observed when moving from 224x224 to 448x448. This improvement is likely due to better preservation of fine-grained visual details in handwritten expressions. However, further increasing the resolution to 768x768 yields only marginal improvements, with performance comparable to that at 448x448. These results suggest that while higher resolutions can capture more detail, the benefits plateau beyond a certain point, making the additional computational cost less justifiable.

\begin{table}
    \centering
    \renewcommand{\arraystretch}{0.8}
  \begin{tabular}{l!{\vrule width \vwidth}cc!{\vrule width \vwidth}cc}
    \toprule
    \multirow{2}{*}{\makecell{\textbf{Image}\\\textbf{Resolutions}}} & \multicolumn{2}{c!{\vrule width \vwidth}}{\textbf{ED}} & \multicolumn{2}{c}{\textbf{EL}} \\
    \cmidrule(lr){2-5}
    & Acc & F1 & Acc & F1 \\
    \midrule
    $ 224\times 224$ & {73.01} & {49.22} & {61.13} & \textbf{58.18} \\
    $ 448\times 448$ & \textbf{75.66} & \textbf{50.42} & \underline{63.87} & \underline{56.57} \\
    $ 768\times 768$ & \underline{75.29} & \underline{50.11} & \textbf{64.58} & {56.50} \\

    \bottomrule
  \end{tabular}
  \caption{\label{tab:ablation_image_size}
    Ablation study results with different image resolutions on the AIHub dataset.}
  
  \vspace{-0.4cm}
\end{table}





%% file: section/5_Conclusion.tex
\section{Conclusion}
We present VEHME, a novel Vision-Language Model tailored for evaluating open-form handwritten mathematical expressions with high accuracy, interpretability, and scalability.
Leveraging a dual-phase training strategy, supervised fine-tuning on expert reasoning traces and reinforcement learning via GRPO, VEHME delivers fine-grained error detection and localization across visually diverse student responses. 
The Expression-Aware Visual Prompting Module further enhances the model’s spatial reasoning capabilities. Experiments on the AIHub and FERMAT datasets show that VEHME consistently outperforms existing open-source models and achieves performance competitive with state-of-the-art proprietary systems. 
These findings underscore VEHME's promise as an effective, open-access solution for scalable, automated assessment in educational settings.

%% file: section/6_Limitations.tex
\section*{Limitations}
While VEHME demonstrates promising results in evaluating handwritten mathematical expressions, several limitations should be acknowledged. First, due to data-sharing restrictions from the source provider (AIHub), the original math problems and student responses cannot be publicly released outside Korea, limiting reproducibility.
Secondly, while VEHME is robust to varied inputs, it may struggle with extremely illegible handwriting, low-quality images, or unconventional notation.

Thirdly, QwQ-32B may propagate biases in our evaluation pipeline, stemming from length-based tendencies or anchoring effects introduced during training and reward design. While output length bias was explicitly controlled, addressing deeper preference and anchoring biases remains an open challenge for future work.

Lastly, the reinforcement learning setup, though effective, can be sensitive to reward design and may introduce training instability. Future work will explore more accessible datasets, improved robustness, and interpretable learning signals.

%% file: section/7_Acknowledgments.tex
\section*{Acknowledgments}
This work was supported by the National Research Foundation of Korea (NRF) grant funded by the Korea government (MSIT) (No. RS-2024–00456247, No. RS-2023–00218913, No. RS-2023-00219959) and by Institute of Information \& communications Technology Planning \& Evaluation (IITP) grant funded by the Korea government(MSIT) (No. RS-2025-25443718, Next-HCAI Project, No.RS-2022-
II220608/2022-0-00608, Artificial intelligence research about multimodal interactions for empathetic conversations with humans, No.IITP-2025-RS-2024-00360227, Leading Generative AI Human Resources Development, No.RS-2025-25442824, AI Star Fellowship Program(Ulsan National Institute of Science and Technology), No.RS-2020-II201336,
Artificial Intelligence graduate school support(UNIST) and No.RS-2019-II191906, Artificial Intelligence Graduate School Program(POSTECH).

This work utilized datasets provided by the Open AI Dataset Project (AIHub, South Korea), which are publicly available through AIHub (www.aihub.or.kr).

%% file: section/X_Appendix.tex
\newpage
\appendix
\section{Appendix}
\label{sec:appendix}

\subsection{Datasets}
\label{sec:datasets}

\begin{figure*}[!b]
  \centering
  \includegraphics[width=0.9\linewidth]{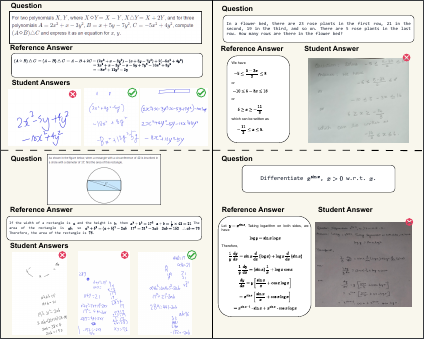}
  \caption {Data samples of AIHub (left) and FERMAT (Right) Datasets.}
  \label{fig:data_sample}
\end{figure*}

Figure~\ref{fig:data_sample} shows representative samples from the AIHub dataset and the FERMAT dataset. Both contain math problems paired with student solutions, but they differ in format and collection style.

\paragraph{AIHub}
comprises 183,085 handwritten student responses corresponding to 30,050 unique problem questions, originally authored in Korean.
To mitigate potential performance degradation due to cross-lingual inputs, we filtered out student answers containing non-English characters. Given that many questions include figures, after translating to English, we further rendered the translated content into LaTeX and converted it into images to ensure compatibility with visual input processing.

\begin{figure}[h]
\centering
  \includegraphics[width=\linewidth]{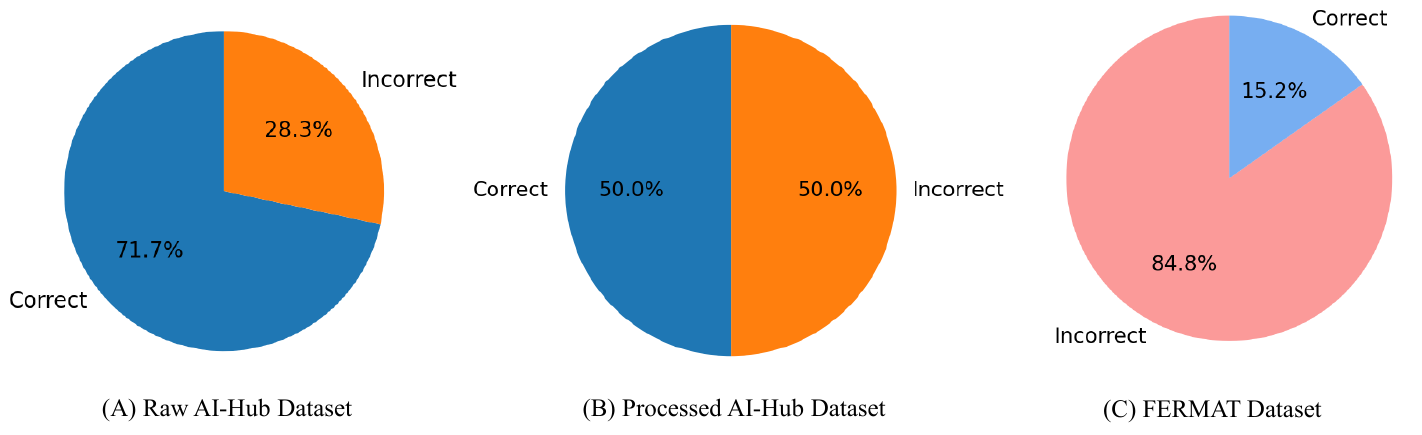}
  \caption {Dataset distribution for correct/incorrect answer.}
  \label{fig:data_distribution}
\end{figure}

As shown in Figure~\ref{fig:data_distribution}, the original AIHub dataset exhibits a significant class imbalance, with approximately $72\%$ of answers being correct and $28\%$ incorrect. To mitigate this skew, we curate the training data as follows: 
\textbf{Step 1:} We discard problems that had only correct or only incorrect student answers, leaving approximately 150,000 problem-answer pairs.
\textbf{Step 2:} For each question, we sample an equal number of correct and incorrect answers to form a balanced training set.
In the end, we obtained a balanced dataset of correct and incorrect solutions from students with 81,394 training samples and 9,062 test samples.

\paragraph{FERMAT} provides a complementary evaluation focused on error analysis, so incorrect solutions significantly outnumber correct ones at an approximate ratio of 85:15, which is different from the balanced distribution in the filtered AIHub dataset. It contains 2,244 manually curated solutions across eight mathematical domains from arithmetic to calculus. Each entry contains: the original question, a handwritten solution image with intentional errors, the gold-standard correct answer, and error detection labels. Due to its small size, we opt not to perform any additional filtering, as doing so could lead to an insufficient amount of data for fine-tuning. We split the dataset into 70\% training and 30\% test sets.

\subsection{Data synthesis from QwQ-32B for SFT}
\label{sec:data-syn}

\begin{figure*}[!b]
\centering
  \includegraphics[width=\linewidth]{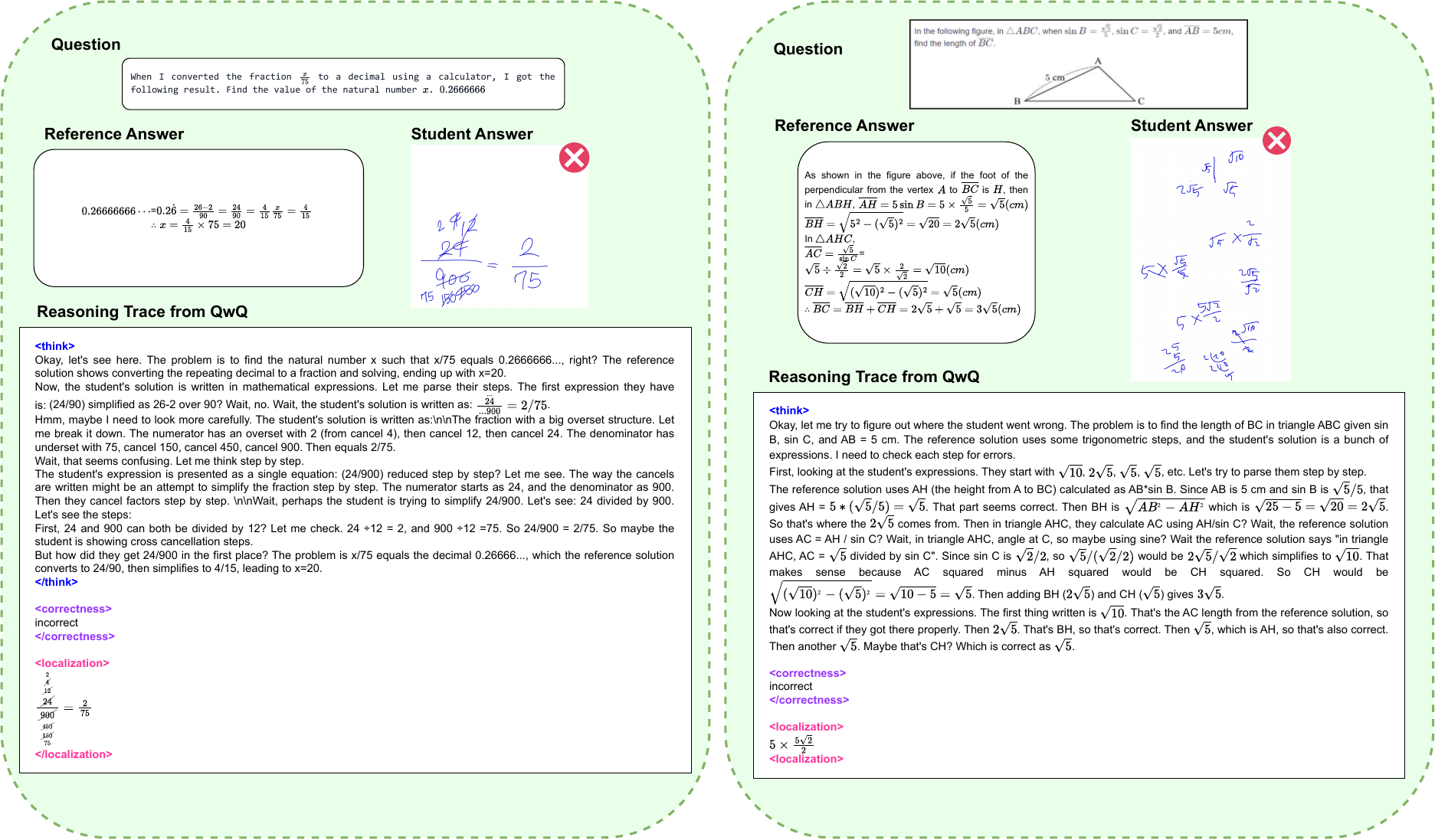}
  \caption {Examples of data synthesized from QwQ-32B for SFT. The model inputs—questions, reference answers, and student responses—are provided in LaTeX format. The figure shown here is for illustrative purposes only; no images are given to the model.}
  \label{fig:data_SFT}
\end{figure*}

Figure~\ref{fig:data_SFT} illustrates some data synthesized from QwQ-32B, showcasing its reasoning process for assessing correctness and localizing errors in students’ handwritten responses.

Please refer to Algorithm~\ref{alg:postprocessing} for a detailed view of the data synthesis procedure.
\begin{algorithm}
\caption{Post-Processing for Generated Data}
\label{alg:postprocessing}
\begin{algorithmic}[1]
\Require Output text $T$, Prompt $P$, Maximum token limit $M$, Grammar constraint $G$, Language model $LLM$
\Ensure Final response with correct format

\If{$\text{TokenCount}(T) < M$}
    \State \Return $T$
\EndIf

\State $T' \gets$ \Call{TruncateLastThought}{$T$}
\State $T'' \gets T' \mathbin{\|} \texttt{"</think>\textbackslash n\textbackslash n<correctness>"}$
\State $P' \gets P \mathbin{\|} T''$
\State Configure $LLM$ with grammar constraint $G$
\State $R \gets$ \Call{LLM.Generate}{$P'$}
\State \Return $P' \mathbin{\|} R$

\end{algorithmic}
\end{algorithm}

The grammar \(G\) used in the decoding process is as follows:

\begin{align*}
    \textit{root} &\;\rightarrow\; \textit{C} \;\texttt{``</correctness>''}\; \textit{L} \\
   \textit{C} &\;\rightarrow\; \texttt{``correct''} \;|\; \texttt{``incorrect''} \\
    \textit{L} &\;\rightarrow\; \texttt{``<localization>''}\; E \;\texttt{``</localization>''}\\
    E &\;\rightarrow\; \texttt{``None''} \;|\; {\Sigma_{\text{math}}}^* \\
    & \;|\; \texttt{``Lack of intermediate steps''} \\
\end{align*}

where \( \Sigma_{\text{math}} \) is the set of valid characters representing mathematical expressions
\[
\Sigma_{\text{math}} = \texttt{[a-zA-Z0-9\textbackslash\{\}\_\^{}\$\textbackslash().,;'"-=<>+|*/]*}
\]


\subsection{Expression-aware Visual Prompting}
\label{sec:evp}

Please refer to Algorithm~\ref{alg:synthetic-canvas} for a formal view of how we synthesize multi-line mathematical expressions.

\begin{algorithm}
\caption{Multi-Line Expression Canvas Synthesis}
\label{alg:synthetic-canvas}
\begin{algorithmic}[1]
\Require Set of \texttt{HMES} files $\mathcal{H}$, padding bound $P$, rotation bound $\theta$
\Ensure Synthesized canvas $\mathcal{C}$ and list of oriented bounding boxes $\mathcal{B}$
\State Render each expression $h_i \in \mathcal{H}$ to image $I_i$
\State Compute canvas size from maximum image width and height
\State Initialize blank canvas $\mathcal{C}$ and vertical offset $y \gets 0$
\ForAll{images $I_i$}
    \State Sample rotation angle $\alpha \sim \mathcal{U}(-\theta, \theta)$
    \State Rotate image with $\alpha$ to obtain $I_i^{\text{rot}}$
    \State Sample horizontal offset 
    $$x \sim \mathcal{U}(0, W - \text{width}{\left(I_i^{\text{rot}}\right)})$$
    \State Compute rotated bounding box corners around image center
    \State Add $(x, y)$-shifted rotated corners to $\mathcal{B}$
    \State Paste $I_i^{\text{rot}}$ onto canvas $\mathcal{C}$ at $(x, y)$
    \State Update $y \gets y + h_i^{\text{rot}} + \text{randint}(-P, P)$
\EndFor
\State \Return Cropped canvas $\mathcal{C}$ and bounding box list $\mathcal{B}$
\end{algorithmic}
\end{algorithm}

\subsection{Implementation Details}
\label{sec:appendix-implementation-details}
To train our model, we apply LoRA \cite{lora} to all MLP layers with a rank of 8 and an $\alpha$-scaling factor of 32. Optimization is performed using the AdamW optimizer \cite{AdamW} with a learning rate of 5e-5, cosine learning rate decay, and a warm-up ratio of 0.05. We set a global batch size of 192 and apply a weight decay of 0.1 to mitigate overfitting. Training is conducted over a single epoch, requiring approximately 6 GPU-days on NVIDIA A100s.

To train our EVPM component, we utilize Yolov11 \cite{Jocher_Ultralytics_YOLO_2023} as the backbone of the bounding box detector. We synthesize $10,000$ training data, and an additional $1000$ validation data. Optimization is performed using the AdamW optimizer \cite{AdamW} with a learning rate of 0.01. We set a global batch size of 16. Training the EVPM is conducted over 200 epochs, requiring only a few hours on a single NVIDIA RTX 3090.

\begin{figure*}
 \centering
  \includegraphics[width=0.75\linewidth]{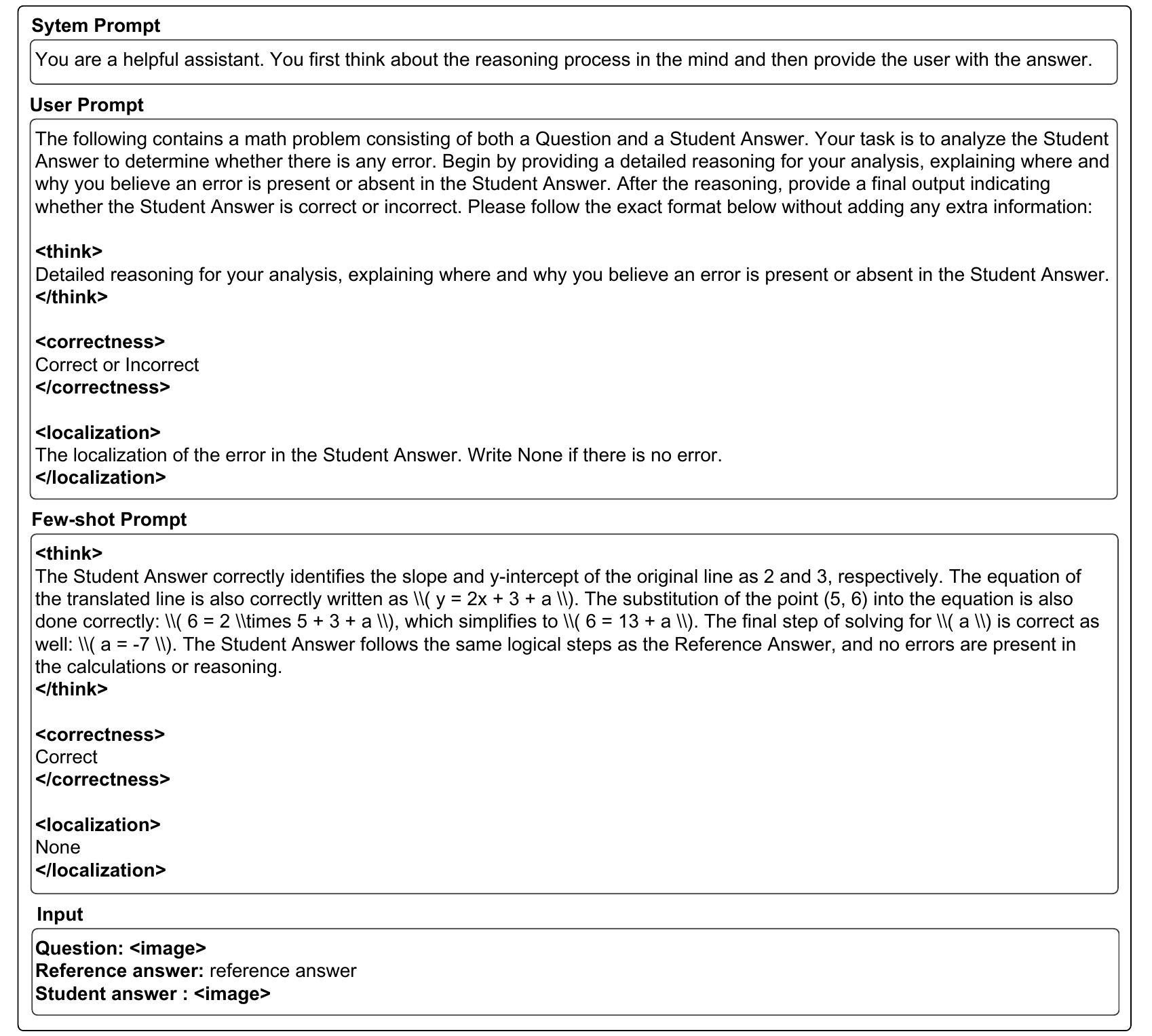}
  \caption {Prompt examples for the inferencing on AIHUB.}
  \label{fig:prompt_aihub}
\end{figure*}

 \begin{figure*}
 \centering
  \includegraphics[width=0.75\linewidth]{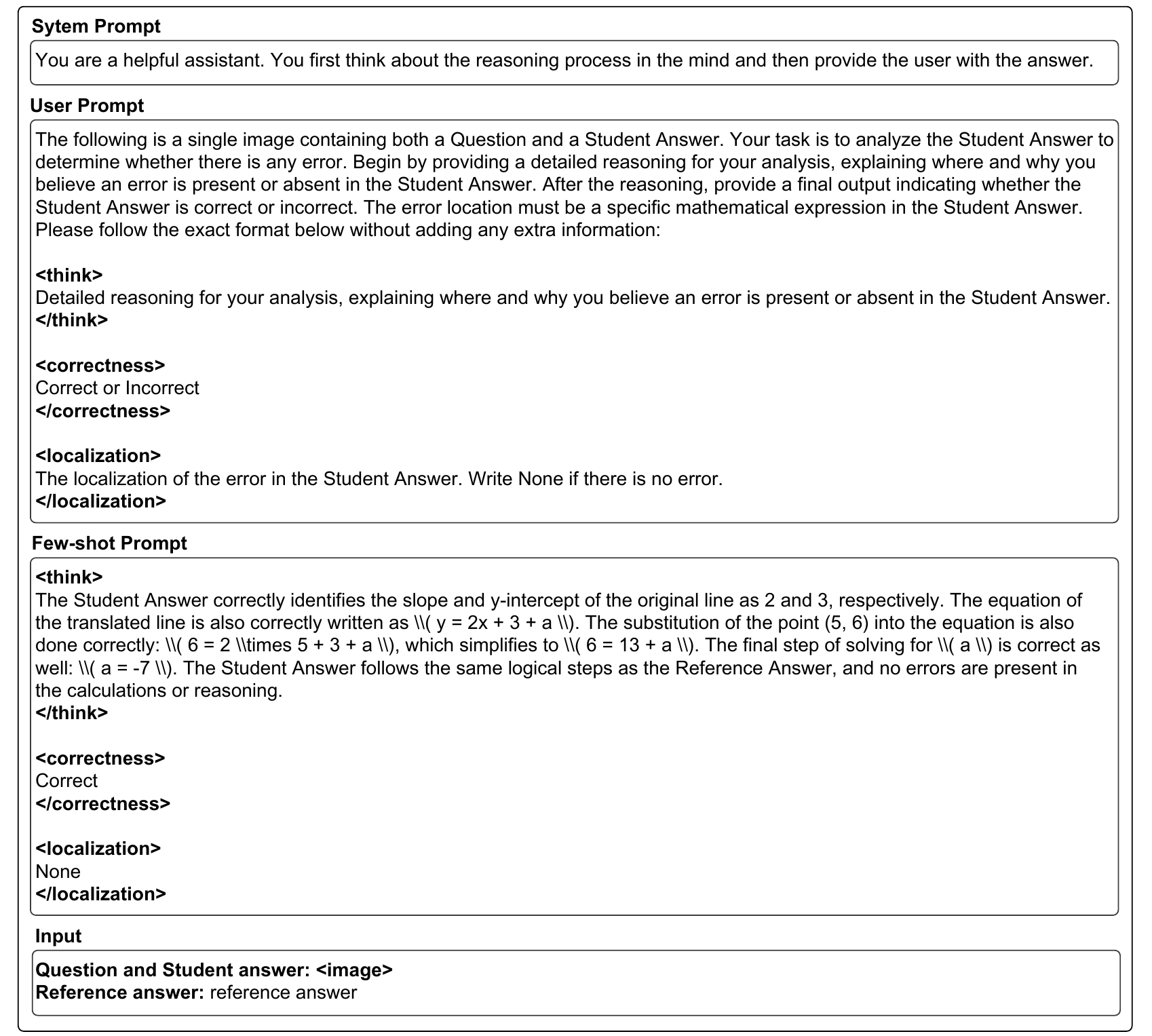}
  \caption {Prompt examples for the inferencing on FERMAT.}
  \label{fig:prompt_fermat}
\end{figure*}

To ensure efficiency during both supervised and reinforcement learning stages, we adopt the \verb|ms-swift| framework \cite{msswift}, which provides optimized memory utilization and high-throughput training for vision-language models. For inference, we employ the \verb|vllm| framework \cite{vllm}, enabling fast, memory-efficient evaluation across all open-source models, including our own. The prompts for inference are presented in Figure~\ref{fig:prompt_aihub} and Figure~\ref{fig:prompt_fermat}.

\subsection{Quantitative Experiment on Different Education Levels}
\label{sec:appendix-exp-education}

\begin{table*}[t]
\centering
\renewcommand{\arraystretch}{0.8}
\scalebox{0.75}{
\begin{tabular}{
    l
    !{\vrule width \vwidth}
    cc|cc
    !{\vrule width \vwidth}
    cc|cc
    !{\vrule width \vwidth}
    cc|cc
}
\toprule
\multirow{2}{*}{\textbf{Models}} 
& \multicolumn{4}{c!{\vrule width \vwidth}}{\textbf{AIHub -- Primary School}} 
& \multicolumn{4}{c!{\vrule width \vwidth}}{\textbf{AIHub -- Middle School}} 
& \multicolumn{4}{c}{\textbf{AIHub -- High School}} \\
\cmidrule(lr){2-13}
& \multicolumn{2}{c|}{\textbf{ED}} & \multicolumn{2}{c!{\vrule width \vwidth}}{\textbf{EL}} 
& \multicolumn{2}{c|}{\textbf{ED}} & \multicolumn{2}{c!{\vrule width \vwidth}}{\textbf{EL}} 
& \multicolumn{2}{c|}{\textbf{ED}} & \multicolumn{2}{c}{\textbf{EL}} \\
& Acc & F1 & Acc & F1 & Acc & F1 & Acc & F1 & Acc & F1 & Acc & F1 \\
\midrule
\rowcolor{gray!20} \multicolumn{13}{c}{Closed-source Models} \\
\midrule

\rowcolor{gray!20} GPT-4o & 74.00 & 49.35 & 64.81 & 64.51 & \textbf{75.24} & \textbf{75.17} & 65.71 & 65.68 & 75.34 & \textbf{75.34} & 66.25 & 66.04 \\
\rowcolor{gray!20} GPT-4o-mini & 58.60 & 34.83 & 40.45 & 38.65 & 56.10 & 31.45 & 37.42 & 34.34 & 55.90 & 31.52 & 39.65 & 35.65 \\
\rowcolor{gray!20} Gemini-2.0-Flash & \textbf{75.59} & \textbf{50.43} & \textbf{67.58} & \textbf{67.52} & 74.98 & 50.01 & \textbf{66.84} & \textbf{66.76} & \textbf{76.40} & 50.95 & \textbf{68.96} & \textbf{68.85} \\
\rowcolor{gray!20} Gemini-2.5-Flash-Preview & 71.20 & 46.97 & 62.94 & 62.67 & 70.91 & 46.86 & 62.56 & 62.35 & 70.70 & 70.23 & 64.99 & 64.82 \\
\midrule
\multicolumn{13}{c}{Open-source Models} \\
\midrule
Phi-4-multimodal-instruct & 38.01 & 28.91 & 35.27 & 32.79 & 39.84 & 30.23 & 36.50 & 34.32 & 35.59 & 27.44 & 36.17 & 34.13 \\
Qwen2.5-VL-7B-Instruct & 46.66 & 32.58 & 39.01 & 38.69 & 47.53 & 33.35 & 39.33 & 39.26 & 48.65 & 33.65 & 37.72 & 37.71 \\
Pixtral-12B & 52.70 & 31.90 & 32.03 & 31.52 & 52.76 & 31.41 & 32.19 & 31.31 & 52.71 & 32.14 & 32.88 & 32.25 \\
Llama-3.2-11B-Vision-Instruct & 19.40 & 18.34 & 44.58 & 33.29 & 19.77 & 18.78 & 44.55 & 33.40 & 18.67 & 18.15 & 46.42 & 32.77 \\
\midrule
\ourModel-Qwen2.5-VL-7B & \textbf{72.10} & \textbf{48.60} & \textbf{60.04} & \textbf{59.85} & \textbf{73.77} & \textbf{49.76} & \textbf{61.50} & \textbf{61.33} & \textbf{73.40} & \textbf{49.35} & \textbf{63.83} & \textbf{63.45} \\
\bottomrule
\end{tabular}
}
\caption{Performance comparisons of state-of-the-art Vision-Language Models on different education levels in the AIHub dataset. The evaluation metrics include Accuracy (Acc) and F1 score (F1). \textbf{ED}: Error Detection, \textbf{EL}: Error Localization. All of the reported results are in percentages ($\%$).}
\label{tab:appendix_education_level_performance}
\end{table*}

\begin{figure*}[!t]
\centering
  \includegraphics[width=0.9\linewidth]{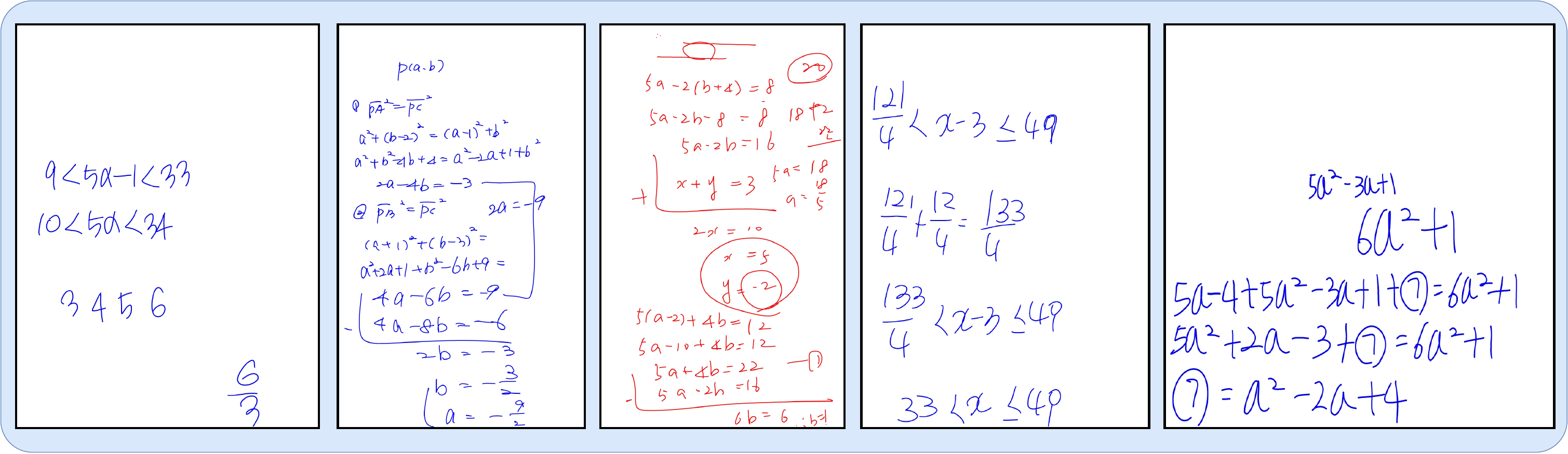}
  \caption {Sample data from the test set of the AIHub dataset.}
  \label{fig:entire_testset}
\end{figure*}

\begin{figure*}[!t]
\centering
  \includegraphics[width=0.9\linewidth]{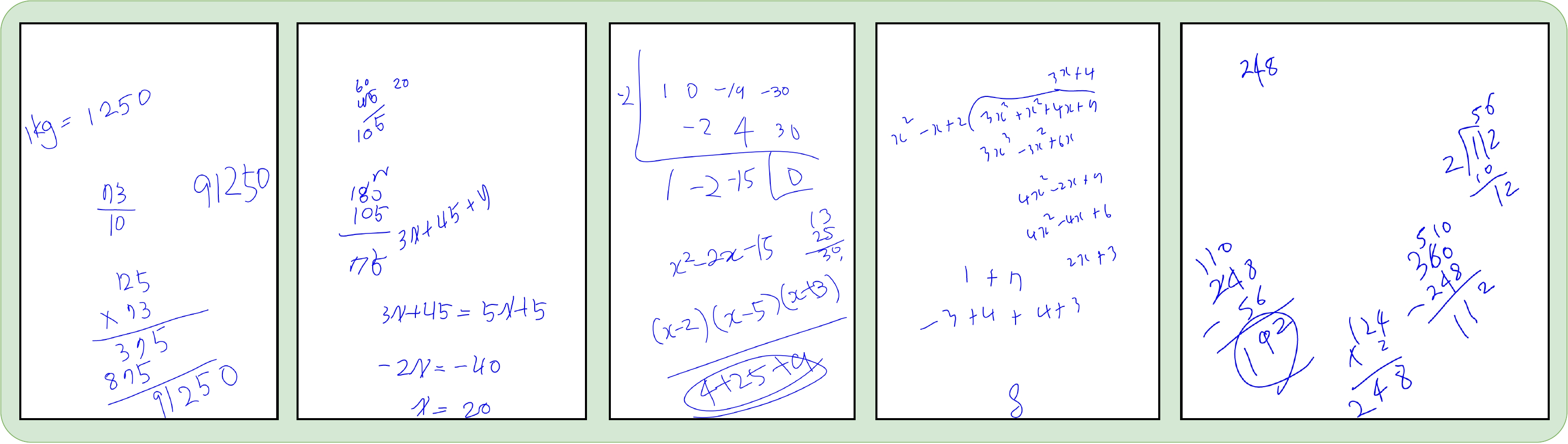}
  \caption {Examples containing heavily rotated mathematical expressions from Challenging Subset of AIHub.}
  \label{fig:challenging}
\end{figure*}

Educational datasets naturally vary in complexity depending on the target grade level: primary school problems tend to be more straightforward and involve simpler arithmetic or reasoning steps, whereas middle and high school problems increasingly incorporate multi-step reasoning, algebraic manipulation, and abstract logic. Since error detection (ED) and error localization (EL) are reasoning-intensive tasks, it is important to evaluate whether models maintain consistent performance across different difficulty levels.
Table~\ref{tab:appendix_education_level_performance} reports results on the AIHub dataset split by education levels (Primary, Middle, and High School). The evaluation again covers Error Detection (ED) and Error Localization (EL). 

Gemini-2.0-Flash emerges as the strongest proprietary baseline across all education levels. It records peak accuracy at the High School level (76.40\% ED / 68.96\% EL), showing consistent gains as the problem difficulty increases. GPT-4o also performs robustly, particularly on Middle School data (75.24\% ED / 65.71\% EL).
The open-source VLMs generally struggle to match closed-source performance, with accuracies ranging between 18–53\%. Pixtral-12B achieves stable ED accuracy ($\approx 52\%$) across levels but remains weak in EL ($\approx 32\%$). Llama-3.2-11B shows relatively higher EL than other open-source peers ($\approx 45\%$), though at the cost of very poor ED performance ($\approx 19\%$). Phi-4 and Qwen2.5-VL-7B-Instruct sit in the middle range, with incremental improvements from Primary to High School, but failing to scale effectively.
\ourModel–Qwen2.5-VL-7B consistently outperforms other open-source competitors across all three education levels. It achieves 72.10\% / 48.60\% (ED) and 60.04\% / 59.85\% (EL) at the Primary School level, climbing to 73.40\% / 49.35\% (ED) and 63.83\% / 63.45\% (EL) at the High School level. This trend demonstrates that \ourModel\ not only generalizes well across increasing difficulty but also narrows the gap with closed-source systems in EL, where precise localization becomes more challenging.

Overall, our findings suggest that performance is largely consistent across grade levels, with no significant degradation or sensitivity to problem complexity. This indicates that the models’ ability to detect and localize errors is not strongly dependent on the education level of the task.

\subsection{Challenging Subset of AIHub}
\label{sec:challenging}
In our analysis, most of the standard dataset has mild rotation $(< 15\degree)$ (as shown in Figure~\ref{fig:entire_testset}), which might explain the relatively small global improvement from EVPM. To better understand EVPM’s contribution, we constructed a ``\textit{Challenging Subset of AIHub}'' by identifying the 100 handwritten data points with the most heavily rotated mathematical expressions (top-100; Mean: $21.81\degree$, Std: $3.40\degree$, Min: $17.56\degree$, Max: $33.20\degree$) as shown in the Figure~\ref{fig:challenging}.

\subsection{Qualitative Result}
\label{sec:appendix-qualitative}
\begin{figure*}[t]
  \includegraphics[width=\linewidth]{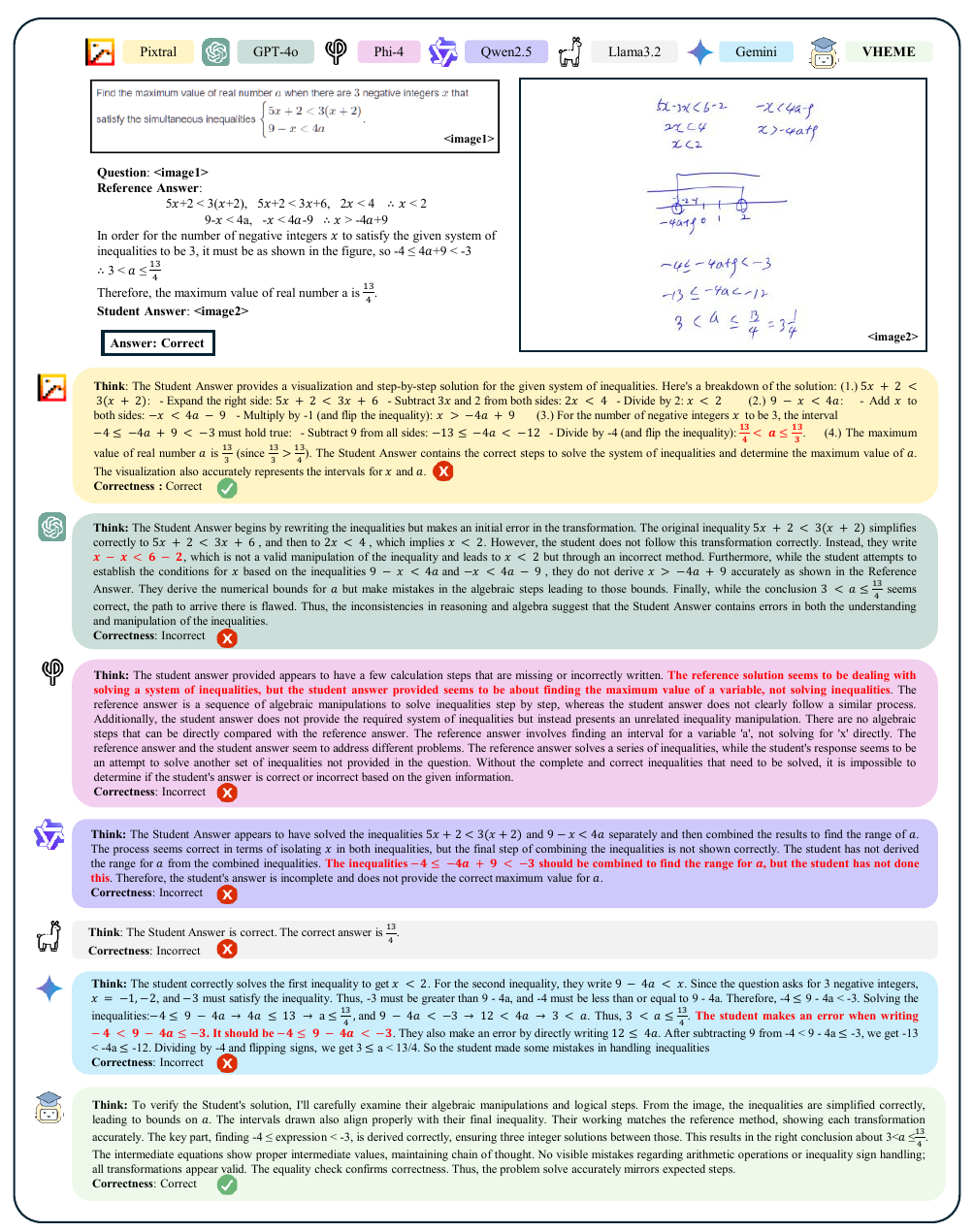}
  \caption {Qualitative example for Error Detection (ED). Among the 7 models, \ourModel\ provides the most correct error detection. The model's error is colored in red. The incorrect response from the models can be categorized into 3 main types: ``\textit{recognition error}'', ``\textit{problem misunderstanding}'', and ``hallucination''. Pixtral, GPT, Qwen2.5, and Gemini all suffer from incorrect OCR of the student's handwritten mathematical expressions. While Pixtral still produces the correct assessment, its thought was wrong due to incorrect OCR. Phi-4 misinterprets the problem as solving the inequality, while the problem is asking to find the maximum value of $a$. Llama3.2 suffers from hallucination where its thought is correct, but the final assessment is \textit{incorrect}. }
  \label{fig:qualitative}
\end{figure*}

\begin{figure*}[t]
  \includegraphics[width=\linewidth]{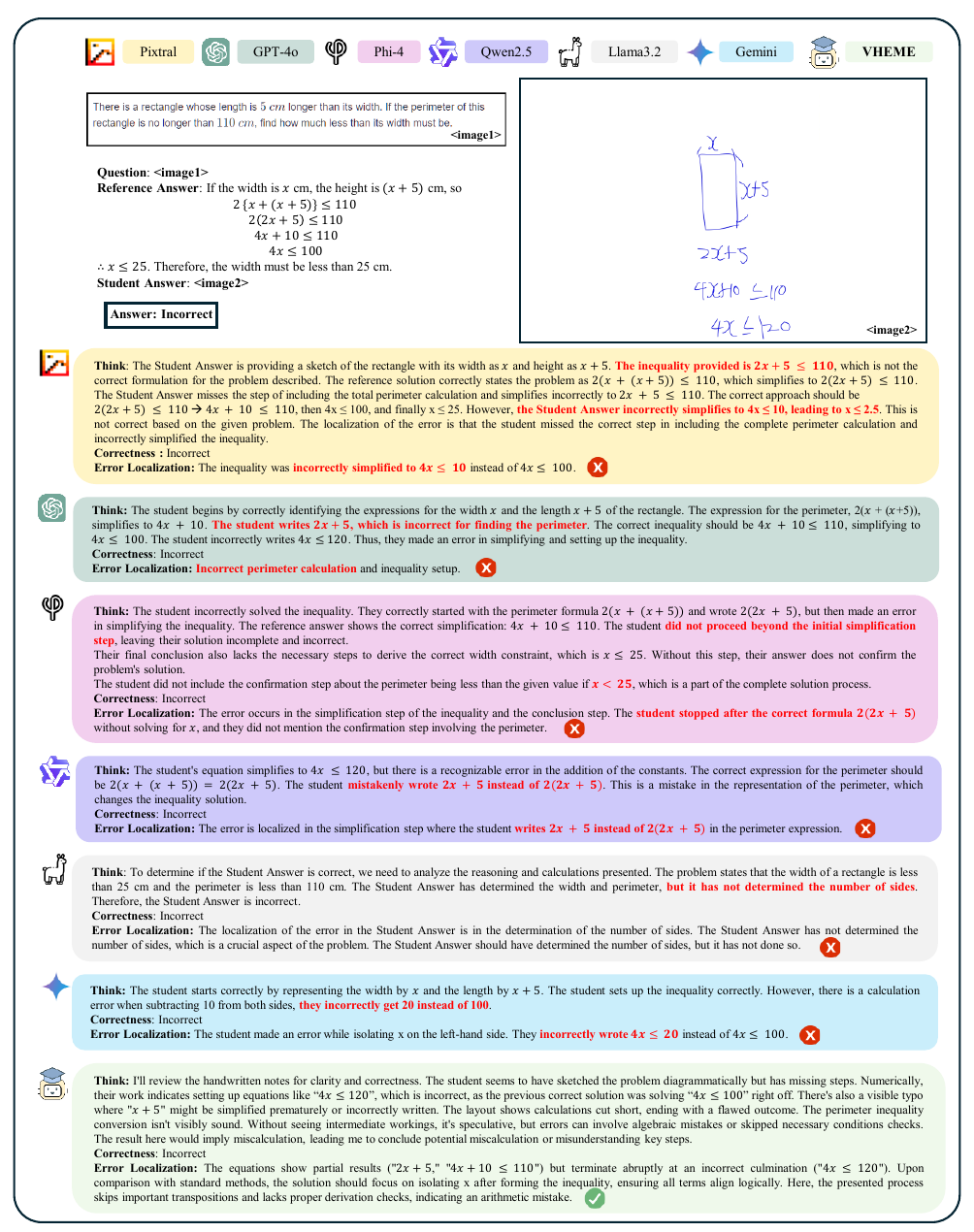}
  \caption {Qualitative example for Error Localization (EL). Among the 7 models, \ourModel\ provides the most correct localization. The model's error is colored in red. The incorrect response from the models can be categorized into 2 main types: ``\textit{recognition error}'' and ``\textit{student's answer misunderstanding}''. Pixtral, Phi-4, and Gemini all suffer from incorrect OCR of the student's handwritten mathematical expressions. Pixtral and Gemini misread the inequality $4x \leq 120$ while Phi-4 cannot read beyond the inequality $2x + 5$. GPT, Qwen2.5, and Llama3.2 misinterpret the student's answer. GPT and Qwen2.5 think that the student writes $2x+5$ as the perimeter, while Llama3.2 thinks that the student must determine the number of sides.
  }
  \label{fig:qualitative_el}
\end{figure*}

We further compare our qualitative results against other baselines, as shown in Figure~\ref{fig:qualitative} and Figure~\ref{fig:qualitative_el}. 

%% file: main.bbl
\begin{thebibliography}{69}
\providecommand{\natexlab}[1]{#1}

\bibitem[{Achiam et~al.(2023)Achiam, Adler, Agarwal, Ahmad, Akkaya, Aleman, Almeida, Altenschmidt, Altman, Anadkat et~al.}]{gpt4}
Josh Achiam, Steven Adler, Sandhini Agarwal, Lama Ahmad, Ilge Akkaya, Florencia~Leoni Aleman, Diogo Almeida, Janko Altenschmidt, Sam Altman, Shyamal Anadkat, and 1 others. 2023.
\newblock Gpt-4 technical report.
\newblock \emph{arXiv preprint arXiv:2303.08774}.

\bibitem[{Bai et~al.(2023)Bai, Bai, Yang, Wang, Tan, Wang, Lin, Zhou, and Zhou}]{bai2023qwenvlversatilevisionlanguagemodel}
Jinze Bai, Shuai Bai, Shusheng Yang, Shijie Wang, Sinan Tan, Peng Wang, Junyang Lin, Chang Zhou, and Jingren Zhou. 2023.
\newblock \href {https://arxiv.org/abs/2308.12966} {Qwen-vl: A versatile vision-language model for understanding, localization, text reading, and beyond}.
\newblock \emph{Preprint}, arXiv:2308.12966.

\bibitem[{Bai et~al.(2025)Bai, Chen, Liu, Wang, Ge, Song, Dang, Wang, Wang, Tang et~al.}]{qwen-vl}
Shuai Bai, Keqin Chen, Xuejing Liu, Jialin Wang, Wenbin Ge, Sibo Song, Kai Dang, Peng Wang, Shijie Wang, Jun Tang, and 1 others. 2025.
\newblock Qwen2. 5-vl technical report.
\newblock \emph{arXiv preprint arXiv:2502.13923}.

\bibitem[{Baral et~al.(2025)Baral, Lucy, Knight, Ng, Soldaini, Heffernan, and Lo}]{DrawEduMath}
Sami Baral, Li~Lucy, Ryan Knight, Alice Ng, Luca Soldaini, Neil Heffernan, and Kyle Lo. 2025.
\newblock Drawedumath: Evaluating vision language models with expert-annotated students' hand-drawn math images.
\newblock In \emph{Proceedings of the 2025 Conference of the North American Chapter of the Association for Computational Linguistics: Human Language Technologies}.

\bibitem[{Callahan et~al.(2021)Callahan, Brant, Mohan, Norman, Arnold, and White}]{callahan2021leveraging}
Mary~E Callahan, Emily~B Brant, Deepika Mohan, Marie~K Norman, Robert~M Arnold, and Douglas~B White. 2021.
\newblock Leveraging technology to overcome the “scalability problem” in communication skills training courses.
\newblock \emph{ATS scholar}, 2(3):327--340.

\bibitem[{Chan and Yeung(1998)}]{chan1998elastic}
Kam-Fai Chan and Dit-Yan Yeung. 1998.
\newblock Elastic structural matching for online handwritten alphanumeric character recognition.
\newblock In \emph{Proceedings. Fourteenth International Conference on Pattern Recognition (Cat. No. 98EX170)}, volume~2, pages 1508--1511. IEEE.

\bibitem[{Chan and Yeung(2000)}]{chan2000mathematical}
Kam-Fai Chan and Dit-Yan Yeung. 2000.
\newblock Mathematical expression recognition: a survey.
\newblock \emph{International Journal on Document Analysis and Recognition}, 3:3--15.

\bibitem[{Chaowicharat and Dejdumrong(2023)}]{chaowicharat2023step}
Ekawat Chaowicharat and Natasha Dejdumrong. 2023.
\newblock A step toward an automatic handwritten homework grading system for mathematics.
\newblock \emph{Information Technology and Control}, 52(1):169--184.

\bibitem[{Chen et~al.(2024{\natexlab{a}})Chen, Chen, Zhang, Wang, Liu, Zhou, Zhang, Wan, Zhou, and Sun}]{chen2024mllm}
Dongping Chen, Ruoxi Chen, Shilin Zhang, Yaochen Wang, Yinuo Liu, Huichi Zhou, Qihui Zhang, Yao Wan, Pan Zhou, and Lichao Sun. 2024{\natexlab{a}}.
\newblock Mllm-as-a-judge: Assessing multimodal llm-as-a-judge with vision-language benchmark.
\newblock In \emph{Forty-first International Conference on Machine Learning}.

\bibitem[{Chen et~al.(2024{\natexlab{b}})Chen, Liao, Li, and Fan}]{chen2024alphamath}
Guoxin Chen, Minpeng Liao, Chengxi Li, and Kai Fan. 2024{\natexlab{b}}.
\newblock \href {https://openreview.net/forum?id=VaXnxQ3UKo} {Alphamath almost zero: Process supervision without process}.
\newblock In \emph{The Thirty-eighth Annual Conference on Neural Information Processing Systems}.

\bibitem[{Chen et~al.(2025{\natexlab{a}})Chen, Tu, Wang, Liu, Tang, Du, Zhou, and Xie}]{Chen2025SFTOR}
Hardy Chen, Haoqin Tu, Fali Wang, Hui Liu, Xianfeng Tang, Xinya Du, Yuyin Zhou, and Cihang Xie. 2025{\natexlab{a}}.
\newblock Sft or rl? an early investigation into training r1-like reasoning large vision-language models.
\newblock \emph{arXiv preprint arXiv:2504.11468}.

\bibitem[{Chen et~al.(2025{\natexlab{b}})Chen, Hu, Zou, Wu, Wang, Hooi, and He}]{chen2025judgelrm}
Nuo Chen, Zhiyuan Hu, Qingyun Zou, Jiaying Wu, Qian Wang, Bryan Hooi, and Bingsheng He. 2025{\natexlab{b}}.
\newblock Judgelrm: Large reasoning models as a judge.
\newblock \emph{ArXiv}, abs/2504.00050.

\bibitem[{Chen et~al.(2024{\natexlab{c}})Chen, Wu, Wang, Su, Chen, Xing, Zhong, Zhang, Zhu, Lu et~al.}]{chen2024internvl}
Zhe Chen, Jiannan Wu, Wenhai Wang, Weijie Su, Guo Chen, Sen Xing, Muyan Zhong, Qinglong Zhang, Xizhou Zhu, Lewei Lu, and 1 others. 2024{\natexlab{c}}.
\newblock Internvl: Scaling up vision foundation models and aligning for generic visual-linguistic tasks.
\newblock In \emph{Proceedings of the IEEE/CVF conference on computer vision and pattern recognition}, pages 24185--24198.

\bibitem[{Deng et~al.(2025)Deng, Bansal, Yin, Peng, Wang, and Chang}]{Deng2025OpenVLThinkerAE}
Yihe Deng, Hritik Bansal, Fan Yin, Nanyun Peng, Wei Wang, and Kai-Wei Chang. 2025.
\newblock Openvlthinker: An early exploration to complex vision-language reasoning via iterative self-improvement.
\newblock \emph{ArXiv}, abs/2503.17352.

\bibitem[{Deng et~al.(2017)Deng, Kanervisto, Ling, and Rush}]{deng2017image}
Yuntian Deng, Anssi Kanervisto, Jeffrey Ling, and Alexander~M Rush. 2017.
\newblock Image-to-markup generation with coarse-to-fine attention.
\newblock In \emph{International Conference on Machine Learning}, pages 980--989. PMLR.

\bibitem[{Geng et~al.(2023)Geng, Josifoski, Peyrard, and West}]{geng2023grammarconstrained}
Saibo Geng, Martin Josifoski, Maxime Peyrard, and Robert West. 2023.
\newblock \href {https://openreview.net/forum?id=KkHY1WGDII} {Grammar-constrained decoding for structured {NLP} tasks without finetuning}.
\newblock In \emph{The 2023 Conference on Empirical Methods in Natural Language Processing}.

\bibitem[{Gowda and Suma(2017)}]{gowda2017comparative}
Ramya~S Gowda and V~Suma. 2017.
\newblock A comparative analysis of traditional education system vs. e-learning.
\newblock In \emph{2017 International conference on innovative mechanisms for industry applications (ICIMIA)}, pages 567--571. IEEE.

\bibitem[{Gu et~al.(2024)Gu, Jiang, Shi, Tan, Zhai, Xu, Li, Shen, Ma, Liu et~al.}]{gu2024survey}
Jiawei Gu, Xuhui Jiang, Zhichao Shi, Hexiang Tan, Xuehao Zhai, Chengjin Xu, Wei Li, Yinghan Shen, Shengjie Ma, Honghao Liu, and 1 others. 2024.
\newblock A survey on llm-as-a-judge.
\newblock \emph{arXiv preprint arXiv:2411.15594}.

\bibitem[{Guo et~al.(2025{\natexlab{a}})Guo, Yang, Zhang, Song, Wang, Zhu, Xu, Zhang, Ma, Bi, Zhang, Yu, Wu, Wu, Gou, Shao, Li, Gao, Liu, Xue, Wang, Wu, Feng, Lu, Zhao, Deng, Ruan, Dai, Chen, Ji, Li, Lin, Dai, Luo, Hao, Chen, Li, Zhang, Xu, Ding, Gao, Qu, Li, Guo, Li, Chen, Yuan, Tu, Qiu, Li, Cai, Ni, Liang, Chen, Dong, Hu, You, Gao, Guan, Huang, Yu, Wang, Zhang, Zhao, Wang, Zhang, Xu, Xia, Zhang, Zhang, Tang, Zhou, Li, Wang, Li, Tian, Huang, Zhang, Wang, Chen, Du, Ge, Zhang, Pan, Wang, Chen, Jin, Chen, Lu, Zhou, Chen, Ye, Wang, Yu, Zhou, Pan, Li, Zhou, Wu, Yun, Pei, Sun, Wang, Zeng, Liu, Liang, Gao, Yu, Zhang, Xiao, An, Liu, Wang, Chen, Nie, Cheng, Liu, Xie, Liu, Yang, Li, Su, Lin, Li, Jin, Shen, Chen, Sun, Wang, Song, Zhou, Wang, Shan, Li, Wang, Wei, Zhang, Xu, Li, Zhao, Sun, Wang, Yu, Zhang, Shi, Xiong, He, Piao, Wang, Tan, Ma, Liu, Guo, Ou, Wang, Gong, Zou, He, Xiong, Luo, You, Liu, Zhou, Zhu, Huang, Li, Zheng, Zhu, Ma, Tang, Zha, Yan, Ren, Ren, Sha, Fu, Xu, Xie, Zhang, Hao, Ma, Yan, Wu, Gu, Zhu, Liu, Li,
  Xie, Song, Pan, Huang, Xu, Zhang, and Zhang}]{deepseek-r1}
Daya Guo, Dejian Yang, Haowei Zhang, Junxiao Song, Peiyi Wang, Qihao Zhu, Runxin Xu, Ruoyu Zhang, Shirong Ma, Xiao Bi, Xiaokang Zhang, Xingkai Yu, Yu~Wu, Z.~F. Wu, Zhibin Gou, Zhihong Shao, Zhuoshu Li, Ziyi Gao, Aixin Liu, and 175 others. 2025{\natexlab{a}}.
\newblock \href {https://doi.org/10.1038/s41586-025-09422-z} {Deepseek-r1 incentivizes reasoning in llms through reinforcement learning}.
\newblock \emph{Nature}, 645(8081):633--638.

\bibitem[{Guo et~al.(2025{\natexlab{b}})Guo, Wang, Yin, Li, and Liu}]{vlpg}
Hong-Yu Guo, Chuang Wang, Fei Yin, Xiao-Hui Li, and Cheng-Lin Liu. 2025{\natexlab{b}}.
\newblock Vision--language pre-training for graph-based handwritten mathematical expression recognition.
\newblock \emph{Pattern Recognition}, 162:111346.

\bibitem[{Guo et~al.(2025{\natexlab{c}})Guo, Yin, Xu, and Liu}]{hievl}
Hong-Yu Guo, Fei Yin, Jian Xu, and Cheng-Lin Liu. 2025{\natexlab{c}}.
\newblock \href {https://doi.org/10.1109/ICASSP49660.2025.10887902} {Hie-vl: A large vision-language model with hierarchical adapter for handwritten mathematical expression recognition}.
\newblock In \emph{ICASSP 2025 - 2025 IEEE International Conference on Acoustics, Speech and Signal Processing (ICASSP)}, pages 1--5.

\bibitem[{He et~al.(2024)He, Wei, Yan, Liu, Wang, Gan, Tu, Liu, Zeng, Wang, Wang, Li, Zhang, Xu, An, Liu, and Zhou}]{skyworkopeno12024}
Jujie He, Tianwen Wei, Rui Yan, Jiacai Liu, Chaojie Wang, Yimeng Gan, Shiwen Tu, Chris~Yuhao Liu, Liang Zeng, Xiaokun Wang, Boyang Wang, Yongcong Li, Fuxiang Zhang, Jiacheng Xu, Bo~An, Yang Liu, and Yahui Zhou. 2024.
\newblock \href {https://huggingface.co/Skywork} {Skywork-o1 open series}.
\newblock \url{https://huggingface.co/Skywork}.

\bibitem[{Hu et~al.(2022)Hu, Shen, Wallis, Allen-Zhu, Li, Wang, Wang, and Chen}]{lora}
Edward~J Hu, Yelong Shen, Phillip Wallis, Zeyuan Allen-Zhu, Yuanzhi Li, Shean Wang, Lu~Wang, and Weizhu Chen. 2022.
\newblock \href {https://openreview.net/forum?id=nZeVKeeFYf9} {Lo{RA}: Low-rank adaptation of large language models}.
\newblock In \emph{International Conference on Learning Representations}.

\bibitem[{Jin et~al.(2025)Jin, Kim, Jung, Kim, Choi, Son, and Kim}]{MathCog}
Hyoungwook Jin, Yoonsu Kim, Dongyun Jung, Seungju Kim, Kiyoon Choi, Jinho Son, and Juho Kim. 2025.
\newblock Investigating large language models in diagnosing students' cognitive skills in math problem-solving.
\newblock \emph{arXiv preprint arXiv:2504.00843}.

\bibitem[{Jocher et~al.(2023)Jocher, Qiu, and Chaurasia}]{Jocher_Ultralytics_YOLO_2023}
Glenn Jocher, Jing Qiu, and Ayush Chaurasia. 2023.
\newblock \href {https://github.com/ultralytics/ultralytics} {Ultralytics yolo (version 8.0.0) [computer software]}.

\bibitem[{Khalifa et~al.(2025)Khalifa, Agarwal, Logeswaran, Kim, Peng, Lee, Lee, and Wang}]{khalifa2025processthink}
Muhammad Khalifa, Rishabh Agarwal, Lajanugen Logeswaran, Jaekyeom Kim, Hao Peng, Moontae Lee, Honglak Lee, and Lu~Wang. 2025.
\newblock Process reward models that think.
\newblock \emph{arXiv preprint arXiv:2504.16828}.

\bibitem[{Kortemeyer et~al.(2024)Kortemeyer, N{\"o}hl, and Onishchuk}]{thermo}
Gerd Kortemeyer, Julian N{\"o}hl, and Daria Onishchuk. 2024.
\newblock Grading assistance for a handwritten thermodynamics exam using artificial intelligence: An exploratory study.
\newblock \emph{Physical Review Physics Education Research}, 20(2):020144.

\bibitem[{Kwon et~al.(2023)Kwon, Li, Zhuang, Sheng, Zheng, Yu, Gonzalez, Zhang, and Stoica}]{vllm}
Woosuk Kwon, Zhuohan Li, Siyuan Zhuang, Ying Sheng, Lianmin Zheng, Cody~Hao Yu, Joseph~E. Gonzalez, Hao Zhang, and Ion Stoica. 2023.
\newblock Efficient memory management for large language model serving with pagedattention.
\newblock In \emph{Proceedings of the ACM SIGOPS 29th Symposium on Operating Systems Principles}.

\bibitem[{Lambert et~al.(2024)Lambert, Pyatkin, Morrison, Miranda, Lin, Chandu, Dziri, Kumar, Zick, Choi et~al.}]{lambert2024rewardbench}
Nathan Lambert, Valentina Pyatkin, Jacob Morrison, LJ~Miranda, Bill~Yuchen Lin, Khyathi Chandu, Nouha Dziri, Sachin Kumar, Tom Zick, Yejin Choi, and 1 others. 2024.
\newblock Rewardbench: Evaluating reward models for language modeling.
\newblock \emph{arXiv preprint arXiv:2403.13787}.

\bibitem[{Laws et~al.(2003)Laws, Howell, and Lindsay}]{laws2003scalability}
R~Dwight Laws, Scott~L Howell, and Nathan~K Lindsay. 2003.
\newblock Scalability in distance education:“can we have our cake and eat it too?”.
\newblock \emph{Online Journal of Distance Learning Administration}, 6(4):80--89.

\bibitem[{Li et~al.(2022)Li, Yuan, Liang, Liu, Ji, Bai, Liu, and Bai}]{li2022counting}
Bohan Li, Ye~Yuan, Dingkang Liang, Xiao Liu, Zhilong Ji, Jinfeng Bai, Wenyu Liu, and Xiang Bai. 2022.
\newblock When counting meets hmer: counting-aware network for handwritten mathematical expression recognition.
\newblock In \emph{European conference on computer vision}, pages 197--214. Springer.

\bibitem[{Li et~al.(2024{\natexlab{a}})Li, Jiang, Zhang, Ren, Liu, Zou, Xu, Li, Yang, Li, Zhang, and Gao}]{Li2024VisualPrompt}
Feng Li, Qing Jiang, Hao Zhang, Tianhe Ren, Shilong Liu, Xueyan Zou, Huaizhe Xu, Hongyang Li, Jianwei Yang, Chunyuan Li, Lei Zhang, and Jianfeng Gao. 2024{\natexlab{a}}.
\newblock Visual in-context prompting.
\newblock In \emph{Proceedings of the IEEE/CVF Conference on Computer Vision and Pattern Recognition (CVPR)}, pages 12861--12871.

\bibitem[{Li et~al.(2025)Li, Zhou, Yang, Zhan, Pan, Ding, Chen, Bo, Lin, and He}]{StepAMC}
Junsong Li, Jie Zhou, Yutao Yang, Bihao Zhan, Qianjun Pan, Yuyang Ding, Qin Chen, Jiang Bo, Xin Lin, and Liang He. 2025.
\newblock Teaching llms for step-level automatic math correction via reinforcement learning.
\newblock \emph{arXiv preprint arXiv:2503.18432}.

\bibitem[{Li et~al.(2019)Li, Yue, Huang, Yang, and Xu}]{Bags}
Xiaoshuo Li, Tiezhu Yue, Xuanping Huang, Zhe Yang, and Gang Xu. 2019.
\newblock Bags: An automatic homework grading system using the pictures taken by smart phones.
\newblock \emph{arXiv preprint arXiv:1906.03767}.

\bibitem[{Li et~al.(2024{\natexlab{b}})Li, Yang, Qi, Jin, Huang, and Ding}]{li2024tree}
Zhe Li, Wentao Yang, Hengnian Qi, Lianwen Jin, Yichao Huang, and Kai Ding. 2024{\natexlab{b}}.
\newblock A tree-based model with branch parallel decoding for handwritten mathematical expression recognition.
\newblock \emph{Pattern Recognition}, 149:110220.

\bibitem[{Lin(2024)}]{lin2024ai}
John~JH Lin. 2024.
\newblock Ai-assisted evaluation of problem-solving performance using eye movement and handwriting.
\newblock \emph{Journal of Research on Technology in Education}, pages 1--25.

\bibitem[{Liu et~al.(2023)Liu, Li, Wu, and Lee}]{liu2023visual}
Haotian Liu, Chunyuan Li, Qingyang Wu, and Yong~Jae Lee. 2023.
\newblock Visual instruction tuning.
\newblock \emph{Advances in neural information processing systems}, 36:34892--34916.

\bibitem[{Liu et~al.(2025)Liu, Sun, Zang, wen Dong, Cao, Duan, Lin, and Wang}]{Liu2025VisualRFTVR}
Ziyu Liu, Zeyi Sun, Yuhang Zang, Xiao wen Dong, Yuhang Cao, Haodong Duan, Dahua Lin, and Jiaqi Wang. 2025.
\newblock Visual-rft: Visual reinforcement fine-tuning.
\newblock \emph{ArXiv}, abs/2503.01785.

\bibitem[{Loshchilov and Hutter(2017)}]{AdamW}
Ilya Loshchilov and Frank Hutter. 2017.
\newblock \href {https://api.semanticscholar.org/CorpusID:53592270} {Decoupled weight decay regularization}.
\newblock In \emph{International Conference on Learning Representations}.

\bibitem[{Mok et~al.(2024)Mok, Akhtar, Clare, Li, Ida, Ross, and Campanelli}]{phys}
Ryan Mok, Faraaz Akhtar, Louis Clare, Christine Li, Jun Ida, Lewis Ross, and Mario Campanelli. 2024.
\newblock Using ai large language models for grading in education: A hands-on test for physics.
\newblock \emph{arXiv preprint arXiv:2411.13685}.

\bibitem[{Nakamoto et~al.(2023)Nakamoto, Flanagan, Dai, Yamauchi, Takami, and Ogata}]{nakamoto2025development}
Ryosuke Nakamoto, Brendan Flanagan, Yiling Dai, Taisei Yamauchi, Kyosuke Takami, and Hiroaki Ogata. 2023.
\newblock Enhancing self-explanation learning through a real-time feedback system: An empirical evaluation study.
\newblock \emph{Sustainability}, 15(21):15577.

\bibitem[{Nath et~al.(2025)Nath, Bathina, Khan, and Khapra}]{fermath}
Oikantik Nath, Hanani Bathina, Mohammed Safi Ur~Rahman Khan, and Mitesh~M Khapra. 2025.
\newblock \href {https://doi.org/10.18653/v1/2025.acl-long.720} {Can vision-language models evaluate handwritten math?}
\newblock In \emph{Proceedings of the 63rd Annual Meeting of the Association for Computational Linguistics (Volume 1: Long Papers)}, pages 14784--14814, Vienna, Austria. Association for Computational Linguistics.

\bibitem[{OpenAI(2024)}]{openai2024reasoning}
OpenAI. 2024.
\newblock \href {https://openai.com/index/learning-to-reason-with-llms/} {Learning to reason with llms}.
\newblock Accessed: 2025-05-19.

\bibitem[{QwenTeam(2025)}]{qwq32b}
QwenTeam. 2025.
\newblock \href {https://qwenlm.github.io/blog/qwq-32b/} {Qwq-32b: Embracing the power of reinforcement learning}.

\bibitem[{Scarlatos et~al.(2025)Scarlatos, Liu, Lee, Baraniuk, and Lan}]{Tutor}
Alexander Scarlatos, Naiming Liu, Jaewook Lee, Richard Baraniuk, and Andrew Lan. 2025.
\newblock \href {https://arxiv.org/abs/2503.06424} {Training llm-based tutors to improve student learning outcomes in dialogues}.
\newblock \emph{Preprint}, arXiv:2503.06424.

\bibitem[{Schulman et~al.(2017)Schulman, Wolski, Dhariwal, Radford, and Klimov}]{schulman2017proximal}
John Schulman, Filip Wolski, Prafulla Dhariwal, Alec Radford, and Oleg Klimov. 2017.
\newblock Proximal policy optimization algorithms.
\newblock \emph{arXiv preprint arXiv:1707.06347}.

\bibitem[{Shao et~al.(2024)Shao, Wang, Zhu, Xu, Song, Bi, Zhang, Zhang, Li, Wu et~al.}]{deepseekmath}
Zhihong Shao, Peiyi Wang, Qihao Zhu, Runxin Xu, Junxiao Song, Xiao Bi, Haowei Zhang, Mingchuan Zhang, YK~Li, Y~Wu, and 1 others. 2024.
\newblock Deepseekmath: Pushing the limits of mathematical reasoning in open language models.
\newblock \emph{arXiv preprint arXiv:2402.03300}.

\bibitem[{Shen et~al.(2025)Shen, Zhang, Zhang, Xu, and Zhao}]{Shen2025VLMR1AS}
Haozhan Shen, Zilun Zhang, Qianqian Zhang, Ruochen Xu, and Tiancheng Zhao. 2025.
\newblock Vlm-r1: A stable and generalizable r1-style large vision-language model. 2025.
\newblock \emph{URL https://arxiv. org/abs/2504.07615}.

\bibitem[{Trung et~al.(2024)Trung, Zhang, Jie, Sun, Jin, and Li}]{trung2024reft}
Luong Trung, Xinbo Zhang, Zhanming Jie, Peng Sun, Xiaoran Jin, and Hang Li. 2024.
\newblock Reft: Reasoning with reinforced fine-tuning.
\newblock In \emph{Proceedings of the 62nd Annual Meeting of the Association for Computational Linguistics (Volume 1: Long Papers)}, pages 7601--7614.

\bibitem[{Truong et~al.(2022)Truong, Nguyen, and Nakagawa}]{truong2022syntactic}
Thanh-Nghia Truong, Cuong~Tuan Nguyen, and Masaki Nakagawa. 2022.
\newblock Syntactic data generation for handwritten mathematical expression recognition.
\newblock \emph{Pattern Recognition Letters}, 153:83--91.

\bibitem[{Wagstaff et~al.(2019)Wagstaff, Lu, and Chen}]{wagstaff2019automatic}
Benjamin Wagstaff, Chiao Lu, and Xiang'Anthony' Chen. 2019.
\newblock Automatic exam grading by a mobile camera: Snap a picture to grade your tests.
\newblock In \emph{Companion Proceedings of the 24th International Conference on Intelligent User Interfaces}, pages 3--4.

\bibitem[{Wang et~al.(2024)Wang, Li, Shao, Xu, Dai, Li, Chen, Wu, and Sui}]{math-shephred}
Peiyi Wang, Lei Li, Zhihong Shao, Runxin Xu, Damai Dai, Yifei Li, Deli Chen, Yu~Wu, and Zhifang Sui. 2024.
\newblock \href {https://doi.org/10.18653/v1/2024.acl-long.510} {Math-shepherd: Verify and reinforce llms step-by-step without human annotations}.
\newblock In \emph{ACL (1)}, pages 9426--9439.

\bibitem[{Wang et~al.(2025)Wang, Gao, Chen, Chen, Zhu, Zhao, Liu, Cao, Ye, Zhu et~al.}]{wang2025visualprm}
Weiyun Wang, Zhangwei Gao, Lianjie Chen, Zhe Chen, Jinguo Zhu, Xiangyu Zhao, Yangzhou Liu, Yue Cao, Shenglong Ye, Xizhou Zhu, and 1 others. 2025.
\newblock Visualprm: An effective process reward model for multimodal reasoning.
\newblock \emph{arXiv preprint arXiv:2503.10291}.

\bibitem[{Wu et~al.(2022)Wu, Du, Li, Zhang, Yang, Ren, and Hu}]{wu2022tdv2}
Changjie Wu, Jun Du, Yunqing Li, Jianshu Zhang, Chen Yang, Bo~Ren, and Yiqing Hu. 2022.
\newblock Tdv2: A novel tree-structured decoder for offline mathematical expression recognition.
\newblock In \emph{Proceedings of the AAAI Conference on Artificial Intelligence}, volume~36, pages 2694--2702.

\bibitem[{Wu et~al.(2020)Wu, Yin, Zhang, Zhang, and Liu}]{wu2020handwritten}
Jin-Wen Wu, Fei Yin, Yan-Ming Zhang, Xu-Yao Zhang, and Cheng-Lin Liu. 2020.
\newblock Handwritten mathematical expression recognition via paired adversarial learning.
\newblock \emph{International Journal of Computer Vision}, 128:2386--2401.

\bibitem[{Wu et~al.(2021)Wu, Yin, Zhang, Zhang, and Liu}]{wu2021graph}
Jin-Wen Wu, Fei Yin, Yan-Ming Zhang, Xu-Yao Zhang, and Cheng-Lin Liu. 2021.
\newblock Graph-to-graph: towards accurate and interpretable online handwritten mathematical expression recognition.
\newblock In \emph{Proceedings of the AAAI Conference on Artificial Intelligence}, volume~35, pages 2925--2933.

\bibitem[{Yang et~al.(2024)Yang, Yang, Zhang, Hui, Zheng, Yu, Li, Liu, Huang, Wei, Lin, Yang, Tu, Zhang, Yang, Yang, Zhou, Lin, Dang, Lu, Bao, Yang, Yu, Li, Xue, Zhang, Zhu, Men, Lin, Li, Tang, Xia, Ren, Ren, Fan, Su, Zhang, Wan, Liu, Cui, Zhang, and Qiu}]{qwen2.5}
An~Yang, Baosong Yang, Beichen Zhang, Binyuan Hui, Bo~Zheng, Bowen Yu, Chengyuan Li, Dayiheng Liu, Fei Huang, Haoran Wei, Huan Lin, Jian Yang, Jianhong Tu, Jianwei Zhang, Jianxin Yang, Jiaxi Yang, Jingren Zhou, Junyang Lin, Kai Dang, and 23 others. 2024.
\newblock Qwen2.5 technical report.
\newblock \emph{arXiv preprint arXiv:2412.15115}.

\bibitem[{Yang et~al.(2023)Yang, Li, Peng, Jin, He, and Yao}]{yang2023read}
Wentao Yang, Zhe Li, Dezhi Peng, Lianwen Jin, Mengchao He, and Cong Yao. 2023.
\newblock Read ten lines at one glance: line-aware semi-autoregressive transformer for multi-line handwritten mathematical expression recognition.
\newblock In \emph{Proceedings of the 31st ACM International Conference on Multimedia}, pages 2066--2077.

\bibitem[{Yeo et~al.(2025)Yeo, Tong, Niu, Neubig, and Yue}]{yeo2025demystifying}
Edward Yeo, Yuxuan Tong, Xinyao Niu, Graham Neubig, and Xiang Yue. 2025.
\newblock \href {https://openreview.net/forum?id=AgtQlhMQ0V} {Demystifying long chain-of-thought reasoning in {LLM}s}.
\newblock In \emph{ICLR 2025 Workshop on Navigating and Addressing Data Problems for Foundation Models}.

\bibitem[{Zhang et~al.(2024{\natexlab{a}})Zhang, Li, Dong, Xu, Zhang, Su, Liu, and Jin}]{zhang2024codedpo}
Kechi Zhang, Ge~Li, Yihong Dong, Jingjing Xu, Jun Zhang, Jing Su, Yongfei Liu, and Zhi Jin. 2024{\natexlab{a}}.
\newblock Codedpo: Aligning code models with self generated and verified source code.
\newblock \emph{arXiv preprint arXiv:2410.05605}.

\bibitem[{Zhang et~al.(2020)Zhang, Yin, Xie, and Lu}]{HmwkCheck}
Lingyu Zhang, Yafeng Yin, Lei Xie, and Sanglu Lu. 2020.
\newblock \href {https://doi.org/10.1145/3410530.3414393} {Hmwkcheck: a homework auto-checking system based on arithmetic operation recognition using smartphones}.
\newblock In \emph{Adjunct Proceedings of the 2020 ACM International Joint Conference on Pervasive and Ubiquitous Computing and Proceedings of the 2020 ACM International Symposium on Wearable Computers}, UbiComp/ISWC '20 Adjunct, page 172–175, New York, NY, USA. Association for Computing Machinery.

\bibitem[{Zhang et~al.(2021)Zhang, Wang, Baraniuk, and Lan}]{zhang2021math}
Mengxue Zhang, Zichao Wang, Richard Baraniuk, and Andrew Lan. 2021.
\newblock Math operation embeddings for open-ended solution analysis and feedback.
\newblock \emph{arXiv preprint arXiv:2104.12047}.

\bibitem[{Zhang et~al.(2025)Zhang, Jiang, Zhang, Lin, and Zhang}]{MathMistakeChecker}
Tianyang Zhang, Zhuoxuan Jiang, Haotian Zhang, Lin Lin, and Shaohua Zhang. 2025.
\newblock Mathmistake checker: A comprehensive demonstration for step-by-step math problem mistake finding by prompt-guided llms.
\newblock In \emph{Proceedings of the AAAI Conference on Artificial Intelligence}, volume~39, pages 29730--29732.

\bibitem[{Zhang et~al.(2024{\natexlab{b}})Zhang, Wu, Yang, Shu, Xiao, Kong, and Sang}]{zhang2024o1}
Yuxiang Zhang, Shangxi Wu, Yuqi Yang, Jiangming Shu, Jinlin Xiao, Chao Kong, and Jitao Sang. 2024{\natexlab{b}}.
\newblock o1-coder: an o1 replication for coding.
\newblock \emph{arXiv preprint arXiv:2412.00154}.

\bibitem[{Zhao and Gao(2022)}]{zhao2022comer}
Wenqi Zhao and Liangcai Gao. 2022.
\newblock Comer: Modeling coverage for transformer-based handwritten mathematical expression recognition.
\newblock In \emph{European conference on computer vision}, pages 392--408. Springer.

\bibitem[{Zhao et~al.(2024)Zhao, Huang, Hu, Wang, Mao, Zhang, Jiang, Wu, Ai, Wang, Zhou, and Chen}]{msswift}
Yuze Zhao, Jintao Huang, Jinghan Hu, Xingjun Wang, Yunlin Mao, Daoze Zhang, Zeyinzi Jiang, Zhikai Wu, Baole Ai, Ang Wang, Wenmeng Zhou, and Yingda Chen. 2024.
\newblock \href {https://arxiv.org/abs/2408.05517} {Swift:a scalable lightweight infrastructure for fine-tuning}.
\newblock \emph{Preprint}, arXiv:2408.05517.

\bibitem[{Zheng et~al.(2024)Zheng, Zhang, Zhang, Lin, Lu, Yu, Liu, Zhou, and Lin}]{processbench}
Chujie Zheng, Zhenru Zhang, Beichen Zhang, Runji Lin, Keming Lu, Bowen Yu, Dayiheng Liu, Jingren Zhou, and Junyang Lin. 2024.
\newblock Processbench: Identifying process errors in mathematical reasoning.
\newblock \emph{arXiv preprint arXiv:2412.06559}.

\bibitem[{Zheng et~al.(2023)Zheng, Chiang, Sheng, Zhuang, Wu, Zhuang, Lin, Li, Li, Xing et~al.}]{zheng2023judging}
Lianmin Zheng, Wei-Lin Chiang, Ying Sheng, Siyuan Zhuang, Zhanghao Wu, Yonghao Zhuang, Zi~Lin, Zhuohan Li, Dacheng Li, Eric Xing, and 1 others. 2023.
\newblock Judging llm-as-a-judge with mt-bench and chatbot arena.
\newblock \emph{Advances in Neural Information Processing Systems}, 36:46595--46623.

\bibitem[{Zhou et~al.(2025)Zhou, Li, Wang, Cheng, Zhou, and Hsieh}]{Zhou2025R1ZerosM}
Hengguang Zhou, Xirui Li, Ruochen Wang, Minhao Cheng, Tianyi Zhou, and Cho-Jui Hsieh. 2025.
\newblock R1-zero's "aha moment" in visual reasoning on a 2b non-sft model.
\newblock \emph{ArXiv}, abs/2503.05132.

\end{thebibliography}
